\documentclass[a4paper,12pt,titlepage,oneside]{kumsthesis}

\usepackage{url}
\usepackage[bottom]{footmisc}
\usepackage[caption=false,font=footnotesize]{subfig}
\usepackage[dvipsnames, table]{xcolor}
\usepackage{amsmath}
\usepackage{amssymb}
\usepackage{amsfonts}
\usepackage{breakcites}
\usepackage{mathtools}
\usepackage{subcaption}
\usepackage{subfig}
\usepackage{adjustbox}
\usepackage{booktabs}
\usepackage{multirow}
\usepackage{pifont}
\usepackage{nccmath}
\usepackage[hidelinks]{hyperref}
\usepackage{makecell, cellspace, caption}
\usepackage{xcolor}

\graphicspath{{./Figures/}}

\usepackage{xspace}
\usepackage{epstopdf}

\newcommand \COMMENT  [1] {}       %

\begin{document}

\Title{Online Long-term Point Tracking in the Foundation Model Era} \Author{Görkay Aydemir} \Year{July 9, 2025}
\TTitle{Temel Modeller Çağında Uzun Vadeli Çevrimiçi Nokta İzleme} \TYear{9 Temmuz 2025}
\Program{Computer Science and Engineering}
\TProgram{Bilgisayar Bilimleri ve M\"{u}hendisli\u{g}i}
\Signature{Assist. Prof. Fatma Güney (Advisor)}
\Signature{Assoc. Prof. Christian Rupprecht}
\Signature{Prof. Ross Goroshin}

\prelimpages
\titlepage
\thesissignaturepage
\dedication{To free and independent academia.}

\abstract{Point tracking aims to identify the same physical point across video frames and serves as a geometry-aware representation of motion. This representation supports a wide range of applications, from robotics to augmented reality, by enabling accurate modeling of dynamic environments.
Most existing long-term tracking approaches operate in an offline setting, where future frames are available to refine predictions and recover from occlusions. However, real-world scenarios often demand online predictions: the model must operate causally, using only current and past frames. This constraint is critical in streaming video and embodied AI, where decisions must be made immediately based on past observations.
Under such constraints, viewpoint invariance becomes essential. Visual foundation models, trained on diverse large-scale datasets, offer the potential for robust geometric representations. While they lack temporal reasoning on their own, they can be integrated into tracking pipelines to enrich spatial features and improve robustness.
In this thesis, we address the problem of long-term point tracking in an online setting, where frames are processed sequentially without access to future information or sliding windows. 
We begin by evaluating the suitability of visual foundation models for this task and find that they can serve as useful initializations and be integrated into tracking pipelines. However, to enable long-term tracking in an online setting, a dedicated design is still required. In particular, maintaining coherence over time in this causal regime requires memory to propagate appearance and context across frames. To address this, we introduce \textbf{Track-On}, a transformer-based model that treats each tracked point as a query and processes video frames one at a time. It predicts correspondences through patch classification followed by local refinement. To ensure temporal consistency, memory modules are incorporated to carry relevant information across frames. Track-On sets a new state of the art across seven public benchmarks, demonstrating the feasibility of long-term tracking without future access.
}

\oz{
Nokta izleme, aynı fiziksel noktanın video çerçeveleri boyunca tanımlanmasını hedefler ve hareketin geometriye duyarlı bir temsili olarak işlev görür. Bu temsil, dinamik ortamların doğru modellenmesini sağlayarak robotikten artırılmış gerçekliğe kadar çeşitli uygulamalara olanak tanır.
Mevcut uzun vadeli izleme yöntemlerinin çoğu, gelecekteki çerçevelerin erişilebilir olduğu çevrimdışı senaryolarda çalışır. Oysa gerçek dünyada, modelin yalnızca geçmiş ve mevcut çerçeveleri kullanarak nedensel şekilde çalışması gereken çevrimiçi tahminler gereklidir.
Bu durumda, bakış açısına karşı değişimsizlik kritik hâle gelir. Büyük ve çeşitli veri kümeleri üzerinde eğitilen görsel temel modeller, sağlam geometrik temsiller sunabilir. Zamanla ilgili çıkarım yetenekleri sınırlı olsa da, bu modeller izleme sistemlerine entegre edilerek mekansal özellikleri zenginleştirebilir.
Bu tezde, geleceğe erişim olmadan çerçevelerin sırayla işlendiği çevrimiçi ortamda uzun vadeli nokta izleme problemini ele alıyoruz. Görsel temel modellerin bu görevde yararlı başlangıç noktaları sunduğunu gösteriyoruz; ancak uzun vadeli izleme için özel bir tasarım gereklidir. Bu amaçla her noktayı bir sorgu olarak işleyen ve video çerçevelerini sırasıyla değerlendiren transformatör tabanlı bir model olan Track-On’u sunuyoruz. Yama sınıflandırması ve yerel iyileştirme ile eşleşmeleri tahmin ederken, bellek modülleriyle zaman boyunca tutarlılığı korur. Track-On, yedi denektaşında geleceğe erişim olmadan uzun vadeli izlemenin mümkün olduğunu göstermektedir.
}
    
\acknowledgments{
I would like to express my deepest gratitude to my advisor, Asst. Professor Fatma Güney, for her continuous support, encouragement, and invaluable guidance throughout the years. I am also sincerely thankful to Assoc. Professor Weidi Xie, whose long-term collaboration and insightful mentorship have played a crucial role in shaping the direction and execution of my research.

I would like to thank Assoc. Professor Christian Rupprecht and Professor Ross Goroshin for kindly agreeing to serve on my thesis committee and for their time and thoughtful evaluation of my work.

I am also grateful to all members of the AVG group for fostering a collaborative research environment, with special thanks to Shadi for his generous help and support.

Beyond the academic journey, I owe everything to my family. Their constant encouragement, quiet sacrifices, and steady belief in me have been a source of strength through every step of this process. I am especially thankful to my mother, Özlem, and my father, Hakan, for their endless support, which has shaped who I am. In a world where everything shifts, their presence has been a quiet constant.

Among my friends, I would like to thank Akın, for being like an older brother, even in academic matters; Mustafa, whose presence transformed my undergraduate years and who became the main figure of my two-year-long Istanbul chapter; and Serdar, with whom I have built not just a friendship but a shared worldview. Together, as “Serdar and Görkay,” we have created a space against the world, where, in the face of absurdity, we embrace the absence of meaning.

Finally, I want to thank Duygu, whose presence brought clarity to a process often clouded by uncertainty. Her patience, warmth, and belief in me never wavered, even during the most difficult periods. I am deeply grateful for her companionship, and for always helping me see beyond the immediate, to what truly matters.
}

\tableofcontents
\listoftables
\listoffigures
\abbreviations{
\begin{tabular}{lp{1cm}l}
    2D & & Two-Dimensional \\
    3D & & Three-Dimensional \\
    AJ & & Average Jaccard \\
    CNN & & Convolutional Neural Network \\
    FoMo & & Foundation Model \\
    IME & & Inference-Time Memory Extension \\
    LoRA & & Low-Rank Adaptation \\
    MLP & & Multi-Layer Perceptron \\
    OA & & Occlusion Accuracy \\
    SD & & Stable Diffusion \\
    TAP & & Tracking-Any-Point \\
    ViT & & Vision Transformer \\
\end{tabular}

}
\textpages

\newcommand{\Perp}{\perp\!\!\! \perp}
\newcommand{\bK}{\mathbf{K}}
\newcommand{\bX}{\mathbf{X}}
\newcommand{\bY}{\mathbf{Y}}
\newcommand{\bk}{\mathbf{k}}
\newcommand{\bx}{\mathbf{x}}
\newcommand{\by}{\mathbf{y}}
\newcommand{\bhy}{\hat{\mathbf{y}}}
\newcommand{\bty}{\tilde{\mathbf{y}}}
\newcommand{\bG}{\mathbf{G}}
\newcommand{\bI}{\mathbf{I}}
\newcommand{\bg}{\mathbf{g}}
\newcommand{\bS}{\mathbf{S}}
\newcommand{\bs}{\mathbf{s}}
\newcommand{\bN}{\mathbf{N}}
\newcommand{\bM}{\mathbf{M}}
\newcommand{\bw}{\mathbf{w}}
\newcommand{\eye}{\mathbf{I}}
\newcommand{\bU}{\mathbf{U}}
\newcommand{\bV}{\mathbf{V}}
\newcommand{\bW}{\mathbf{W}}
\newcommand{\bn}{\mathbf{n}}
\newcommand{\bv}{\mathbf{v}}
\newcommand{\bwv}{\mathbf{wv}}
\newcommand{\bq}{\mathbf{q}}
\newcommand{\bR}{\mathbf{R}}
\newcommand{\bi}{\mathbf{i}}
\newcommand{\bj}{\mathbf{j}}
\newcommand{\bp}{\mathbf{p}}
\newcommand{\bt}{\mathbf{t}}
\newcommand{\bJ}{\mathbf{J}}
\newcommand{\bu}{\mathbf{u}}
\newcommand{\bB}{\mathbf{B}}
\newcommand{\bD}{\mathbf{D}}
\newcommand{\bz}{\mathbf{z}}
\newcommand{\bP}{\mathbf{P}}
\newcommand{\bC}{\mathbf{C}}
\newcommand{\bA}{\mathbf{A}}
\newcommand{\bZ}{\mathbf{Z}}
\newcommand{\bff}{\mathbf{f}}
\newcommand{\bF}{\mathbf{F}}
\newcommand{\bo}{\mathbf{o}}
\newcommand{\bO}{\mathbf{O}}
\newcommand{\bc}{\mathbf{c}}
\newcommand{\bm}{\mathbf{m}}
\newcommand{\bT}{\mathbf{T}}
\newcommand{\bQ}{\mathbf{Q}}
\newcommand{\bL}{\mathbf{L}}
\newcommand{\bl}{\mathbf{l}}
\newcommand{\ba}{\mathbf{a}}
\newcommand{\bE}{\mathbf{E}}
\newcommand{\bH}{\mathbf{H}}
\newcommand{\bd}{\mathbf{d}}
\newcommand{\br}{\mathbf{r}}
\newcommand{\be}{\mathbf{e}}
\newcommand{\bb}{\mathbf{b}}
\newcommand{\bh}{\mathbf{h}}
\newcommand{\bhh}{\hat{\mathbf{h}}}
\newcommand{\btheta}{\boldsymbol{\theta}}
\newcommand{\bTheta}{\boldsymbol{\Theta}}
\newcommand{\bpi}{\boldsymbol{\pi}}
\newcommand{\bPi}{\boldsymbol{\Pi}}
\newcommand{\bphi}{\boldsymbol{\phi}}
\newcommand{\bPhi}{\boldsymbol{\Phi}}
\newcommand{\bmu}{\boldsymbol{\mu}}
\newcommand{\bSigma}{\boldsymbol{\Sigma}}
\newcommand{\bGamma}{\boldsymbol{\Gamma}}
\newcommand{\bbeta}{\boldsymbol{\beta}}
\newcommand{\bomega}{\boldsymbol{\omega}}
\newcommand{\blambda}{\boldsymbol{\lambda}}
\newcommand{\bLambda}{\boldsymbol{\Lambda}}
\newcommand{\bkappa}{\boldsymbol{\kappa}}
\newcommand{\btau}{\boldsymbol{\tau}}
\newcommand{\balpha}{\boldsymbol{\alpha}}
\newcommand{\nR}{\mathbb{R}}
\newcommand{\nN}{\mathbb{N}}
\newcommand{\nL}{\mathbb{L}}
\newcommand{\cN}{\mathcal{N}}
\newcommand{\cM}{\mathcal{M}}
\newcommand{\cR}{\mathcal{R}}
\newcommand{\cB}{\mathcal{B}}
\newcommand{\cL}{\mathcal{L}}
\newcommand{\cH}{\mathcal{H}}
\newcommand{\cS}{\mathcal{S}}
\newcommand{\cT}{\mathcal{T}}
\newcommand{\cO}{\mathcal{O}}
\newcommand{\cC}{\mathcal{C}}
\newcommand{\cP}{\mathcal{P}}
\newcommand{\cE}{\mathcal{E}}
\newcommand{\cI}{\mathcal{I}}
\newcommand{\cF}{\mathcal{F}}
\newcommand{\cK}{\mathcal{K}}
\newcommand{\cY}{\mathcal{Y}}
\newcommand{\cX}{\mathcal{X}}
\newcommand{\cQ}{\mathcal{Q}}
\newcommand{\cV}{\mathcal{V}}
\def\bgamma{\boldsymbol\gamma}

\newcommand{\specialcell}[2][c]{%
  \begin{tabular}[#1]{@{}c@{}}#2\end{tabular}}

\newcommand{\figref}[1]{Figure~\ref{#1}}
\newcommand{\secref}[1]{Section~\ref{#1}}
\newcommand{\algref}[1]{Algorithm~\ref{#1}}
\newcommand{\eqnref}[1]{Equation~\eqref{#1}}
\newcommand{\tabref}[1]{Table~\ref{#1}}
\newcommand{\mychapref}[1]{Chapter~\ref{#1}}

\newcommand{\rulesep}{\unskip\ \vrule\ }

\newcommand{\KLD}[2]{D_{\mathrm{KL}} \left( \left. \left. #1 \right|\right| #2 \right) }

\renewcommand{\b}{\ensuremath{\mathbf}}

\def\mc{\mathcal}
\def\mb{\mathbf}

\newcommand{\T}{^{\raisemath{-1pt}{\mathsf{T}}}}

\makeatletter
\DeclareRobustCommand\onedot{\futurelet\@let@token\@onedot}
\def\@onedot{\ifx\@let@token.\else.\null\fi\xspace}
\def\eg{e.g\onedot} \def\Eg{E.g\onedot}
\def\ie{i.e\onedot} \def\Ie{I.e\onedot}
\def\cf{cf\onedot} \def\Cf{Cf\onedot}
\def\etc{etc\onedot} \def\vs{vs\onedot}
\def\wrt{wrt\onedot}
\def\dof{d.o.f\onedot}
\def\etal{et~al\onedot} \def\iid{i.i.d\onedot}
\def\Fig{Fig\onedot} \def\Eqn{Eqn\onedot} \def\Sec{Sec\onedot} \def\Alg{Alg\onedot}
\makeatother

\renewcommand{\b}{\ensuremath{\mathbf}}

\def\mc{\mathcal}
\def\mb{\mathbf}

\makeatletter
\DeclareRobustCommand\onedot{\futurelet\@let@token\@onedot}
\def\@onedot{\ifx\@let@token.\else.\null\fi\xspace}
\def\eg{{\em e.g}\onedot}
\def\Eg{E.g\onedot}
\def\ie{{\em i.e}\onedot} \def\Ie{I.e\onedot}
\def\cf{cf\onedot} \def\Cf{Cf\onedot}
\def\etc{{\em etc}\onedot} \def\vs{vs\onedot}
\def\wrt{wrt\onedot}
\def\dof{d.o.f\onedot}
\def\etal{et~al\onedot} \def\iid{i.i.d\onedot}
\def\Fig{Fig\onedot} \def\Eqn{Eqn\onedot} \def\Sec{Sec\onedot} \def\Alg{Alg\onedot}
\makeatother

\newcommand{\mytexttilde}{\raisebox{0.5ex}{\texttildelow}}

\newcommand*\rot{\rotatebox{90}}
\newcommand{\boldparagraph}[1]{\noindent{\bf #1:} }
\newcommand{\boldquestion}[1]{\noindent{\bf #1} }

\newcommand{\weidi}[1]{ \noindent {\color{magenta} {\bf Weidi:} {#1}} }
\newcommand{\ftm}[1]{ \noindent {\color{cyan} {\bf Fatma:} {#1}} }
\newcommand{\ga}[1]{ \noindent {\color{blue} {\bf Gorkay:} {#1}} }

\newcommand{\cmark}{\ding{51}}%
\newcommand{\xmark}{\ding{55}}%

\newcommand{\deltaavg}{$\delta^{x}_{avg}$}

\newcommand{\up}{$\uparrow$}
\newcommand{\down}{$\downarrow$}

\chapter{Introduction}
\label{chapter:intro}
Motion estimation is one of the core challenges in computer vision, with applications spanning video compression~\cite{Jasinschi1998JFI}, video stabilization~\cite{Battiato2007ICIAP, Lee2009ICCV}, and augmented reality~\cite{Marchand2015VCG}. The objective is to track physical points across video frames accurately. A widely used solution is optical flow, which estimates pixel-level correspondences between adjacent frames. In principle, long-term motion estimation can be achieved by chaining together these frame-by-frame estimations.

Recent advances in optical flow techniques, such as PWC-Net~\cite{Sun2018CVPR} and RAFT~\cite{Teed2020ECCV}, have significantly improved short-term accuracy. However, chaining flow predictions over long time horizons remains problematic due to error accumulation and challenges in handling occlusions. 

\begin{figure}
    \centering
    \includegraphics[width=\linewidth]{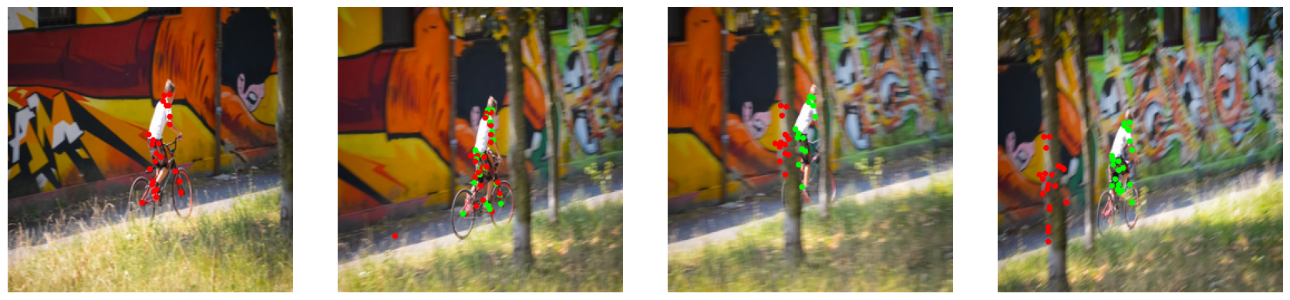}
    \caption{
        \textbf{Tracking with Optical Flow.} When short-term optical flow predictions are chained to achieve long-term tracking, they fail to recover from occlusions. In this example, all predictions ({\color{red} red}) collapse onto the tree trunk, while the correct locations ({\color{green} green}) are on the cycling child.
    } 
    \label{fig:optical_flow}
\end{figure}

For example, in~\figref{fig:optical_flow}, long-term correspondences are obtained by chaining short-term optical flow predictions. While this works in unobstructed regions, the chain fails to recover occluded points: all predictions are mistakenly aligned with the tree trunk, even though the actual points lie on the cycling child.

To address this,~\cite{Sand2008IJCV} proposed pixel tracking, a paradigm shift that focuses on tracking individual points through time. This idea, revisited in PIPs~\cite{Harley2022ECCV} and TAPIR~\cite{Doersch2023ICCV}, leverages deep features and cost volumes to maintain correspondence, a task commonly referred to as tracking-any-point (TAP), or simply, point tracking.

Point tracking extends the correspondence problem into a long-term setting~\cite{Harley2022ECCV, Doersch2022NeurIPS}, where the objective is to estimate the 2D projection, \ie the location in each video frame, of the same physical point throughout the video. This requires overcoming significant challenges such as appearance changes, occlusions, and complex motion. The task demands a high level of geometric awareness, as it goes beyond pairwise frame matching and requires consistent reasoning over time. Importantly, long-term point tracking plays a fundamental role in many real-world applications: in robotics, it enables precise object manipulation~\cite{Vecerik2023ICRA}; in augmented reality, it allows stable anchoring of virtual content; in autonomous navigation, it supports robust motion estimation and scene understanding; and in 3D vision, it benefits tasks like structure-from-motion and SLAM by providing reliable feature trajectories. 

A key requirement in this setting is viewpoint invariance: the ability to recognize and follow a physical point despite changes in camera pose, perspective, and lighting. As the visual appearance of a point may vary drastically under motion, trackers must rely on geometric features rather than raw appearance. This makes geometry-aware feature representations, particularly those that are invariant to viewpoint transformations, highly desirable.

\section{Foundation Models for Point Tracking}

Visual foundation models (FoMos), pretrained on large-scale and diverse datasets, have demonstrated strong generalization across a variety of downstream tasks, including classification~\cite{Radford2021ICML}, segmentation~\cite{Wang2023CVPR}, and object localization~\cite{Melas2022CVPR}. Their ability to extract semantically and geometrically meaningful features makes them attractive candidates for point tracking. Specifically, their potential for viewpoint-invariant representations may help mitigate appearance variations and occlusion-induced drift in long-term tracking.

Matching and correspondence estimation provide a natural way to probe the geometric capacity of such models. If a model can reliably associate pixels corresponding to the same physical point across views or time, this indicates a strong inductive bias toward geometric consistency. Indeed, several recent studies~\cite{Tang2023NeurIPS, Zhang2024CVPR, El2024CVPR} have explored the correspondence capabilities of foundation models, primarily focusing on two-view matching.

However, point tracking goes beyond static matching: it extends correspondence estimation into a dynamic and long-term context. Here, the model must handle ongoing changes in appearance, scene dynamics, and visibility. Successfully tracking a single point through an entire video requires consistent and temporally coherent representations, something that simple two-view matching does not guarantee.

This raises a natural question: do the powerful representations learned by visual foundation models contain sufficient geometric and temporal structure to support long-term point tracking, without any task-specific supervision? Unlike models trained explicitly for correspondence, FoMos are optimized for general visual understanding, often via contrastive or masked pretraining objectives. Despite this, their dense features often encode surprising spatial regularities and object-level consistency.

To answer this, we evaluate whether these models can serve as a basis for long-term point tracking. Specifically, we investigate their capabilities in a zero-shot setting without any task-specific training, and analyze how well they can be adapted when supervision is available. This allows us to probe the geometric inductive biases of foundation models and understand their potential and limitations as generic backbones for temporally consistent tracking.

In the first part of this thesis, we systematically assess whether foundation models can be repurposed for point tracking, and to what extent they can represent and maintain long-term geometric correspondences.

While foundation models offer strong geometric representations, they must be integrated into full tracking pipelines to realize their potential in real-world settings. In particular, long-term point tracking requires not only spatially rich features but also temporal reasoning, a component missing from most frozen backbones.

\section{Online Point Tracking}

\begin{figure}[t]
    \centering
    \includegraphics[width=\linewidth]{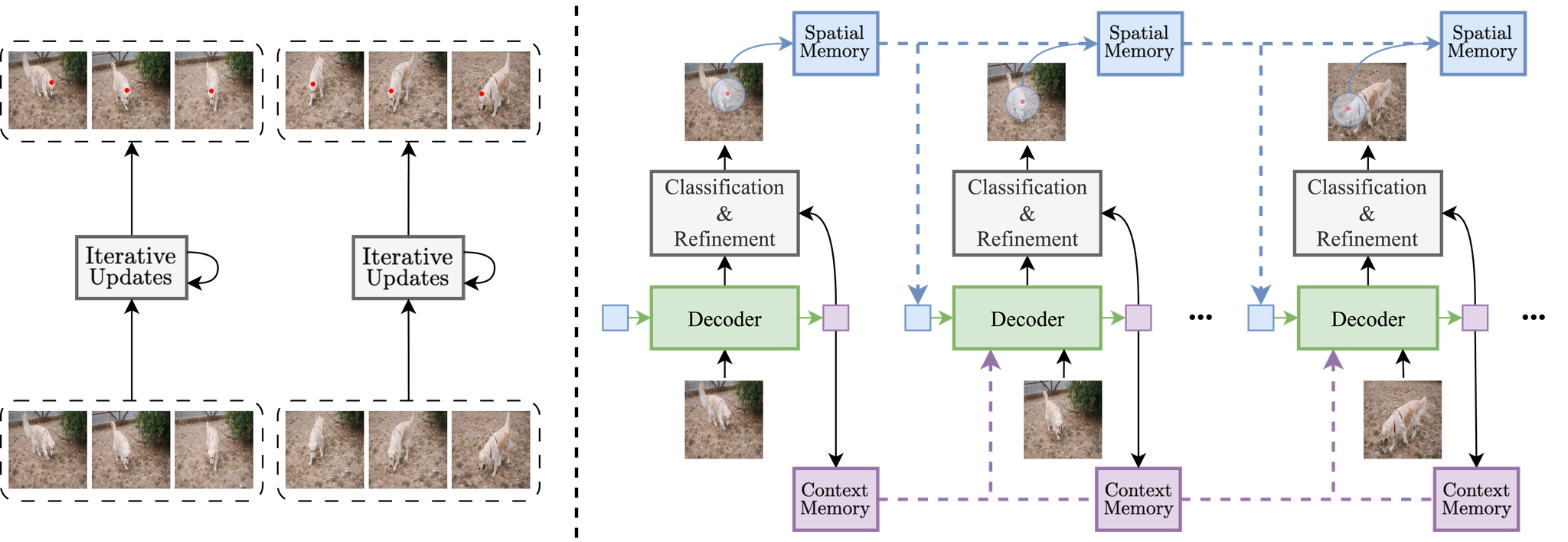}
    \caption{
    \textbf{Offline vs. Online Point Tracking.} We propose an online model, tracking points frame-by-frame (\textbf{right}), unlike the dominant offline paradigm where models require access to all frames within a sliding window or the entire video %
    (\textbf{left}).
    In contrast, our approach allows for frame-by-frame tracking in videos of any length.
    To capture temporal information, we introduce two memory modules: {\color{NavyBlue} spatial memory}, which tracks changes in the target point, and {\color{Purple} context memory}, which stores broader contextual information from previous states of the point.} 
    \label{fig:teaser}
\end{figure}

Even when correspondence-aware features are used within existing methods, already achieving strong performance, most models remain fundamentally offline. They process entire videos or large frame windows, leveraging both past and future information to refine predictions (\figref{fig:teaser}, left). While effective, this design limits their applicability in real-time scenarios~\cite{Karaev2024ECCV, Harley2022ECCV}, where predictions must be made immediately without access to future frames. For instance, in streaming video analysis or robotics, the environment evolves in response to the agent’s current actions, requiring the model to operate causally and produce predictions on the fly.

Furthermore, many offline methods rely on full spatiotemporal attention across video sequences, leading to significant memory overhead and poor scalability to longer sequences. These limitations motivate the need for online models that can reason about motion and visibility in a streaming setting, without sacrificing tracking accuracy or robustness.

Prior work on online point tracking remains limited. Online-TAPIR~\cite{Vecerik2023ICRA} adapts an offline model~(TAPIR~\cite{Doersch2023ICCV}) by applying a causal attention mask and retraining it for sequential inference. While this enables causal operation, the model architecture remains offline in its original design, it does not incorporate components specifically tailored for online tracking, such as mechanisms for maintaining and updating memory over time. In contrast, tasks like video object segmentation~\cite{Ravi2025ICLR} and action recognition~\cite{Xu2021NeurIPS} have demonstrated the effectiveness of purpose-built memory modules in capturing long-term temporal dependencies. These insights suggest that achieving reliable online point tracking requires architectural choices that go beyond causal masking, including dedicated memory structures that support temporal continuity and robust appearance adaptation.

In the second part of this thesis, we address the challenge of long-term point tracking in an online setting (\figref{fig:teaser}, right), where the model processes video frames sequentially without access to future frames. We propose a simple transformer-based model, \textbf{Track-On}, in which points of interest are treated as queries in a transformer decoder that attends to the current frame to update their representations. Rather than relying on full temporal modeling, our approach maintains temporal continuity through two specialized memory modules: \textit{spatial memory} and \textit{context memory}. This enables the model to perform reliable long-term tracking while avoiding the high computational and memory costs associated with dense video-wide attention.

Specifically, spatial and context memory play distinct but complementary roles. The former reduces drift by updating the query representation using recent visual content around the tracked point. This ensures that the query reflects the most recent appearance, rather than the initial one. The latter maintains a broader summary of the track’s history by storing past query embeddings, capturing long-term evolution and occlusion events. Together, these modules allow the model to reason over time while operating causally and efficiently.

At training time, the queries identify the most likely location by computing embedding similarity with patches in the current frame and are supervised using similarity-based classification, akin to contrastive learning. The prediction is further refined through a local offset regression to determine precise coordinates. We show through extensive experiments that our patch-classification and refinement approach offers a compelling alternative to dominant iterative update schemes~\cite{Karaev2024ECCV, Harley2022ECCV, Doersch2023ICCV}. Our model sets a new state-of-the-art among online methods and achieves performance on par with or surpassing offline models across seven benchmark datasets, including TAP-Vid~\cite{Doersch2022NeurIPS}.

\section{Findings}

In this thesis, we demonstrated that visual foundation models can provide powerful initializations for long-term point tracking. Although not explicitly trained for correspondence, their representations offer strong spatial regularities that can be leveraged for tracking tasks. Our systematic evaluation showed that, with minimal adaptation, these models encode geometric structure that supports temporally consistent predictions.

Building on this insight, we introduced \textbf{Track-On}, a simple yet effective transformer-based model for online point tracking. Unlike dominant iterative update approaches optimized with regression, Track-On formulates point tracking as a patch classification problem. Our results show that this formulation, combined with a lightweight refinement step via local offset prediction, can achieve high-precision correspondences while avoiding the complexity of iterative updates.

A central component of our model is its memory design. Our spatial memory module reduces the effects of drift by continually updating the query representation based on recent visual context. In parallel, the context memory provides a longer-term view by accumulating query embeddings over time, enabling the model to reason about appearance changes and occlusions. Importantly, the memory size can be fixed during training and flexibly extended at inference, allowing the model to generalize to arbitrarily long videos.

Track-On achieves state-of-the-art performance among online trackers and matches or surpasses offline baselines across a diverse set of benchmarks: from high-quality object-centric datasets like DAVIS, to large-scale internet videos such as Kinetics, to synthetic environments like Dynamic Replica, and very-long sequences in Point Odyssey. Crucially, our model remains highly efficient: each frame is processed independently with a single backbone forward pass, while relevant past information is compactly stored and accessed through our memory design.

These results underscore the potential of online point tracking as a scalable and practical alternative to traditional offline approaches. By combining causal processing with effective memory mechanisms, Track-On closes the gap between real-time operation and long-term correspondence. We believe that online point tracking will play an increasingly central role in dynamic vision systems, enabling deployment in real-world settings such as robotics, augmented reality, and embodied AI.

\section{Thesis Outline and Contributions}

This thesis is organized as follows: Chapter~\ref{chapter:rw} reviews related work on optical flow, long-term point tracking, visual foundation models, and online tracking architectures with memory-based designs.

Chapter~\ref{chapter:eval_fomo} presents an analysis~\cite{Aydemir2024ECCVW}\footnote{Project page: {\color{magenta} \url{https://kuis-ai.github.io/fomo_pt}}} of visual foundation models in the context of long-term point tracking under zero-shot settings. We show that certain models offer strong geometric representations that serve as effective initializations for temporal correspondence tasks, and that lightweight adaptation can match or exceed the performance of fully supervised baselines.

Chapter~\ref{chapter:track_on} introduces \textbf{Track-On}~\cite{Aydemir2025ICLR}\footnote{Track-On project page: {\color{magenta} \url{https://kuis-ai.github.io/track_on}}}, a transformer-based model for online point tracking that treats target points as queries and performs correspondence estimation via patch classification and refinement. Through extensive experiments and ablations, we demonstrate that Track-On achieves state-of-the-art performance among online trackers and performs competitively with offline methods.

Finally, Chapter~\ref{chapter:conc} concludes the thesis and outlines several directions for future work to further advance the capabilities of long-term and online tracking.

In addition to the two main contributions presented in this thesis, two other research projects were completed during the same period but are not included in the core chapters:  
(i) \textbf{SOLV}~\cite{Aydemir2024NeurIPS}\footnote{SOLV project page: {\color{magenta} \url{https://kuis-ai.github.io/solv}}}, the first fully unsupervised method for segmenting multiple objects in real-world videos using an object-centric approach. By combining a novel masking strategy with slot merging based on similarity, SOLV effectively segments diverse object classes in challenging YouTube videos.  
(ii) \textbf{ADAPT}~\cite{Aydemir2023ICCV}\footnote{ADAPT project page: {\color{magenta} \url{https://kuis-ai.github.io/adapt}}}, a method for predicting the trajectories of all agents in complex traffic scenes. Leveraging dynamic weight learning and an adaptive prediction head, ADAPT achieves superior performance and efficiency compared to prior multi-agent forecasting methods.

Taken together, these works reflect a unified research centered on modeling temporal dynamics, spanning object-centric reasoning, multi-agent prediction, and point-level tracking. This thesis contributes to the broader goal of building models that understand and act upon time-varying information across diverse domains, from autonomous driving to general-purpose video understanding.

\chapter{Related Work} %
\label{chapter:rw}
\section{Optical Flow} 
\begin{figure}
    \centering
    \includegraphics[width=1\linewidth]{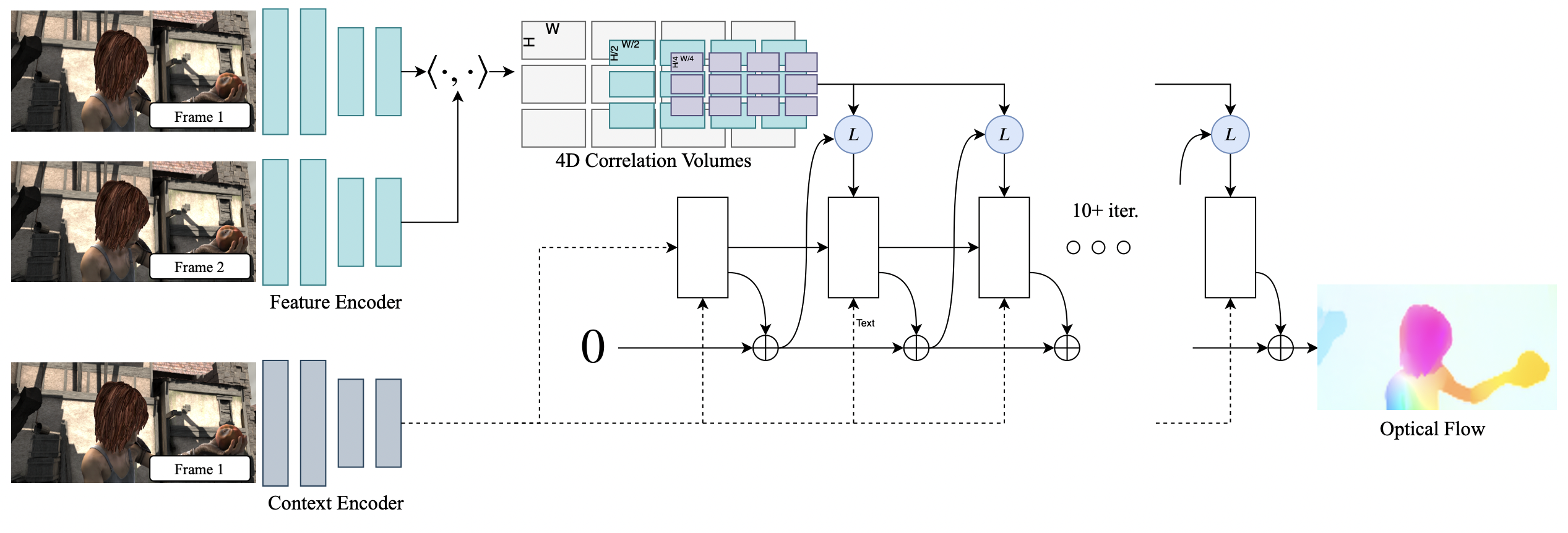}
    \caption{Iterative \textit{update operation} introduced in \textbf{RAFT}~\cite{Teed2020ECCV}.}
    \label{fig:raft}
\end{figure}

Optical flow estimation has long been a fundamental problem in computer vision, addressing the pixel-wise correspondence between consecutive frames. Classical methods~\cite{Horn1981AI, Black1993ICCV, Bruhn2005IJCV} laid the foundation through energy minimization and variational approaches. With the advent of deep learning, FlowNet~\cite{Dosovitskiy2015ICCV} introduced the first end-to-end trainable neural architectures for optical flow. Subsequently, methods incorporating explicit cost volume computations, such as DCFlow~\cite{Xu2017CVPR} and PWC-Net~\cite{Sun2018CVPR}, significantly improved local correspondence matching. RAFT~\cite{Teed2020ECCV}, refined this further by iteratively updating flow predictions via recurrent processing of cost volumes (\figref{fig:raft}), achieving notable accuracy and inspiring subsequent methods~\cite{Zhang2021ICCV, Shi2023CVPR}. Despite these advancements, optical flow methods inherently operate between pairs of consecutive frames, limiting their ability to handle long-term occlusions and significant appearance changes~\cite{Neoral2024WACV, Doersch2022NeurIPS}. Recent approaches aiming to address these limitations typically integrate additional temporal context or explicit tracking mechanisms~\cite{Neoral2024WACV}.

\section{Vision Foundation Models} 
In recent years, the availability of large-scale datasets and increased computing power has led to the development of deep learning models that can handle various visual tasks. These models are trained on large amounts of data through methods like generative objectives~\cite{Rombach2022CVPR}, self-supervision~\cite{Caron2021ICCV, Oquab2024TMLR}, or supervised learning on extensive labeled datasets~\cite{Kirillov2023ICCV}. The flexibility of these models allows them to perform different  tasks, such as video object segmentation~\cite{Wang2023CVPR, Wang2023ARXIV}, estimating correspondence~\cite{Hedlin2024NeurIPS}, object-centric learning~\cite{Aydemir2024NeurIPS}, discovering parts~\cite{Amir2021ARXIV}, autonomous driving~\cite{Barin2024ECCVW}, and open-vocabulary segmentation~\cite{Xu2023CVPR}.

\section{Point Tracking}
\begin{figure}
    \centering
    \includegraphics[width=1\linewidth]{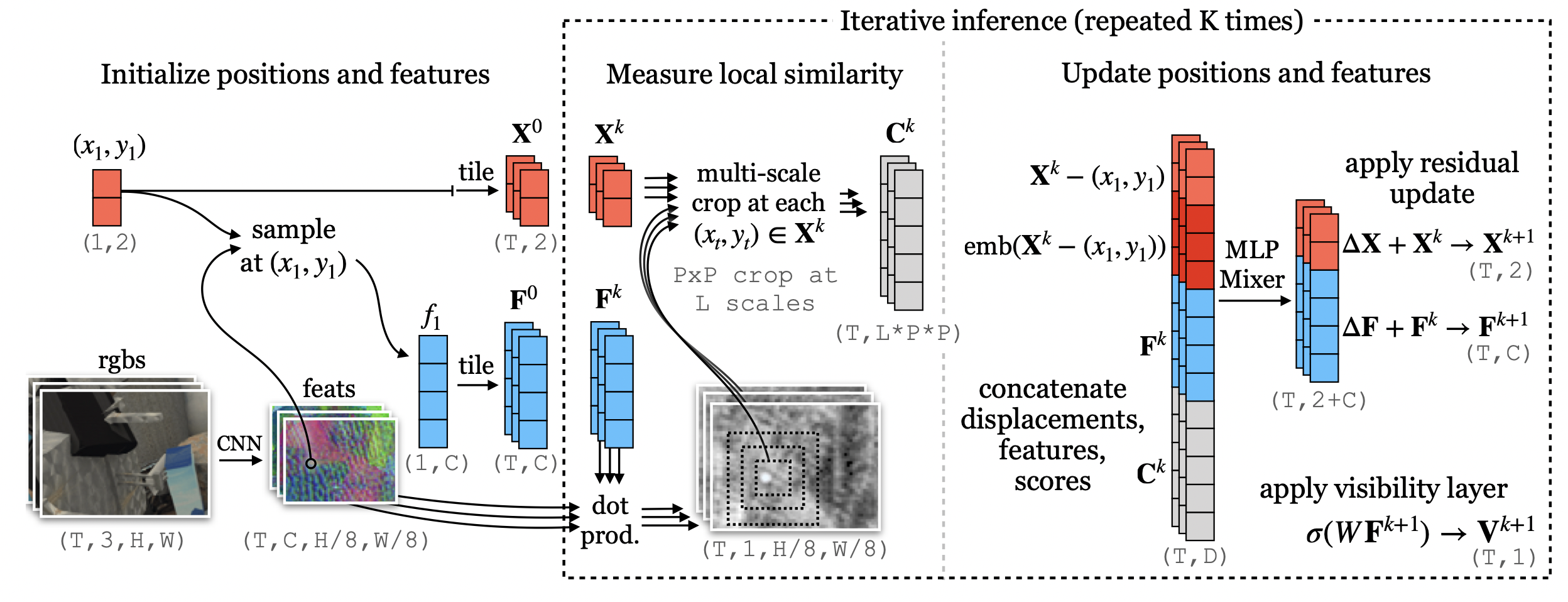}
    \caption{Overview of \textbf{PIPs}~\cite{Harley2022ECCV}, iteratively updating the track feature and correspondence prediction in the temporal window.}
    \label{fig:pips}
\end{figure}

Point tracking, presents significant challenges, particularly for long-term tracking where maintaining consistent tracking through occlusions is difficult. PIPs~\cite{Harley2022ECCV} was one of the first approaches to address this by predicting motion through iterative updates within temporal windows, as illustrated in~\figref{fig:pips}. TAP-Vid~\cite{Doersch2022NeurIPS} initiated a benchmark for evaluation. TAPIR~\cite{Doersch2023ICCV} improved upon PIPs by refining initialization and incorporating depthwise convolutions to enhance temporal accuracy. BootsTAPIR~\cite{Doersch2024ARXIV} further advanced TAPIR by utilizing student-teacher distillation on a large corpus of real-world videos.
In contrast, CoTracker~\cite{Karaev2024ECCV} introduced a novel approach by jointly tracking multiple points, exploiting spatial correlations between points via factorized transformers. Differently, TAPTR~\cite{Li2024ECCV} adopted a design inspired by DETR~\cite{Carion2020ECCV, Zhu2021ICLR}, drawing parallels between object detection and point tracking. DINO-Tracker~\cite{Tumanyan2024ECCV} took a different route, using DINO as a foundation for test-time optimization. TAPTRv2~\cite{Li2024NeurIPS}, the successor to TAPTR, builds on its predecessor by incorporating offsets predicted by the deformable attention module. While these models calculate point-to-region similarity for correlation, LocoTrack~\cite{Cho2024ECCV} introduced a region-to-region similarity approach to address ambiguities in matching. Recently, CoTracker3~\cite{Karaev2024ARXIV} combined the region-to-region similarity method from LocoTrack with the original CoTracker architecture and utilized pseudo-labeled real-world data during training to further enhance performance.

However, all these models are designed for offline tracking, assuming access to all frames within a sliding window~\cite{Karaev2024ECCV} or the entire video~\cite{Doersch2023ICCV, Doersch2024ARXIV}. Conversely, MFT~\cite{Neoral2024WACV}, which extends optical flow to long-term scenarios, can be adapted for online point tracking tasks, although it does not belong to the point tracking family. Among point tracking approaches, models with online variants~\cite{Doersch2024ARXIV, Doersch2023ICCV} are re-trained with a temporally causal mask to process frames sequentially on a frame-by-frame basis, despite being originally designed for offline tracking. In contrast, we explicitly focus on online point tracking by design, enabled by novel memory modules to capture temporal information. Additionally, many of these models use a regression objective, originally developed for optical flow~\cite{Teed2020ECCV}, while we introduce a new paradigm based on patch classification and refinement.

Another line of research, orthogonal to ours, explores leveraging scene geometry for point tracking. SpatialTracker~\cite{Xiao2024CVPR} extends CoTracker to the 3D domain by tracking points in three-dimensional space, while OmniMotion~\cite{Wang2023ICCV} employs test-time optimization to learn a canonical representation of the scene. Concurrent work DynOMO~\cite{Seidenschwarz2025THREEDV} also uses test-time optimization, utilizing Gaussian splats for online point tracking.

\section{Causal Processing in Videos} 
Online, or temporally causal models rely solely on current and past frames without assuming access to future frames. This is in contrast to current practice in point tracking with clip-based models, processing frames together. Causal models are particularly advantageous for streaming video understanding~\cite{Yang2022CVPRb, Zhou2024CVPR}, embodied perception~\cite{Yao2019IROS}, and processing long videos~\cite{Zhang2024ARXIV, Xu2021NeurIPS}, as they process frames sequentially, making them well-suited for activation caching. Due to its potential, online processing has been studied across various tasks in computer vision, such as pose estimation~\cite{Fan2021ICCV, Nie2019ICCV}, action detection~\cite{Xu2019ICCV, De2016ECCV, Kondratyuk2021CVPR, Eun2020CVPR, Yang2022CVPRa, Wang2021ICCV, Zhao2022ECCV, Xu2021NeurIPS, Chen2022CVPR}, temporal action localization~\cite{Buch2017CVPR, Singh2017ICCV}, object tracking~\cite{He2018CVPR, Wang2020ECCV}, video captioning~\cite{Zhou2024CVPR}, and video object segmentation~\cite{Cheng2022ECCV, Liang2020NeurIPS}.

\begin{figure}[t]
    \centering
    \includegraphics[width=1\linewidth]{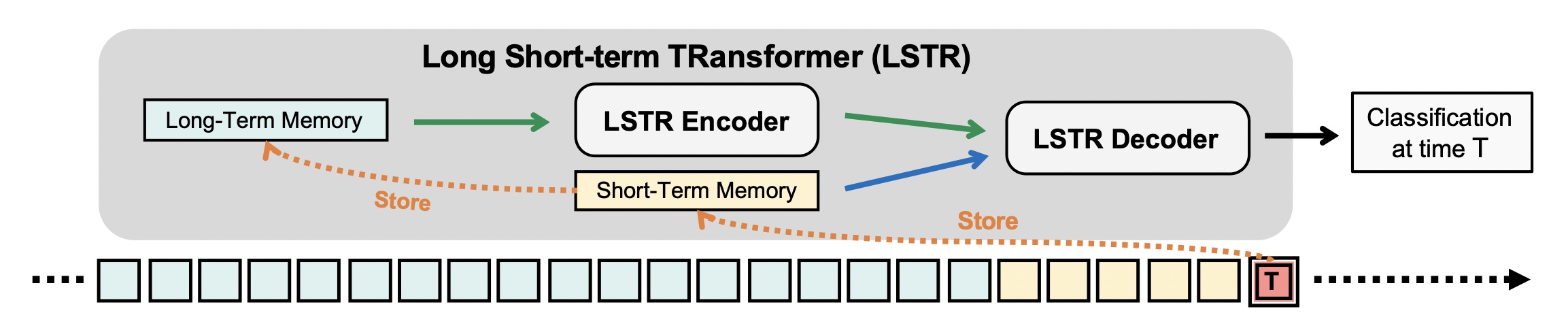}
    \caption{\textbf{LSTR}~\cite{Xu2021NeurIPS} maintains causal representations by combining short-term and long-term memory modules to store and update information in a streaming video setting.}
    \label{fig:lstr}
\end{figure}

In causal models, information from past context is commonly propagated using either sequential models~\cite{De2016ECCV}, which are inherently causal, or transformers with causal attention masks~\cite{Wang2021ICCV}. However, these models often struggle to retain information over long contexts or face expanded memory requirements when handling extended past contexts. To address this, some approaches introduce memory modules for more effective and efficient handling of complex tasks. For example, LSTR~\cite{Xu2021NeurIPS} separates past context as long-term and short-term memories for action detection (\figref{fig:lstr}), while XMem~\cite{Cheng2022ECCV} incorporates a sensory memory module for fine-grained information in video object segmentation. Long-term memory-based modeling is also applied beyond video understanding~\cite{Balazevic2024ICML}, including tasks like long-sequence text processing and video question answering~\cite{Zhang2021ICML}. 
We also employ an attention-based memory mechanism, which is specialized in point tracking with two types of memory; one focusing on spatial local regions around points, and another on broader context. 

\chapter{Can Visual Foundation Models Achieve Long-term Point Tracking?}
\label{chapter:eval_fomo}
Vision foundation models (FoMos) trained on large-scale data have demonstrated strong generalization across diverse visual tasks~\cite{Radford2021ICML, Oquab2024TMLR, Rombach2022CVPR}, yet their capacity to support long-term correspondence estimation remains underexplored. In this chapter, we investigate whether these models can be effectively repurposed for long-term point tracking. Our goal is to assess their geometric awareness and determine whether tracking can emerge through simple mechanisms like feature similarity or lightweight adaptation.

To this end, we formulate a point tracking pipeline that leverages correlation maps computed from FoMo features, and we evaluate models under three distinct regimes: (i) a \textit{zero-shot} setting, where we use the frozen model to retrieve the most similar location based on cosine similarity~\cite{El2024CVPR, Zhan2023ARXIV}; (ii) a \textit{probing} setup, where low-capacity convolutional heads are trained to decode point positions and occlusions from the correlation map~\cite{Doersch2022NeurIPS}; and (iii) an \textit{adaptation} setup, where we apply Low-Rank Adaptation (LoRA)\cite{Hu2022ICLR} to fine-tune attention layers.

\section{Methodology}
In this section, we first explain the models we consider~(\secref{sec:eval_fomo:method:models}) and how we leverage them for point tracking under different settings~(\secref{sec:eval_fomo:method:eval}).

\subsection{Models}
\label{sec:eval_fomo:method:models}
We consider a range of visual models, each trained with different objectives on diverse datasets, inspired by previous work~\cite{El2024CVPR}. Specifically, we evaluate models trained with self-supervision: Masked Autoencoders (MAE)~\cite{He2022CVPR}, DINO~\cite{Caron2021ICCV}, DINOv2~\cite{Oquab2024TMLR}, and DINOv2 with registers (DINOv2-Reg)~\cite{Darcet2023ARXIV}; language supervision: CLIP~\cite{Radford2021ICML}; image generation: Stable Diffusion (SD)~\cite{Rombach2022CVPR}; and direct supervision for classification and segmentation respectively: DeIT III~\cite{Touvron2022ECCV} and Segment Anything (SAM)~\cite{Kirillov2023ICCV}. For each model, we use the available pre-trained checkpoints.

All models, except SD, utilize Vision Transformer (ViT) architectures, while SD employs a U-Net based model. For the ViTs, we use the Base architecture (ViT-B) with a patch size of either 14 (for the DINO family) or 16 (for the remaining ViTs), unless explicitly stated otherwise. We use the output of the last block as our feature map, discarding the \texttt{[CLS]} token.

For SD, we follow the approach outlined in DIFT~\cite{Tang2023NeurIPS} and use the outputs from an intermediate layer of the U-Net encoder. Specifically, we utilize the outputs of the upsample block at $n = 2$, after noising the input frame with a time step of $t = 51$, corresponding to $1/8^{\text{th}}$ of the input image resolution. To reduce the stochasticity of the noising process, we calculate the average feature over 8 runs, each with different noisy versions of the same input image.

\subsection{FoMos for Point Tracking}
\label{sec:eval_fomo:method:eval}

\begin{figure}[t]
    \centering
    \includegraphics[width=.8\linewidth, trim={1.25cm 0.75cm 1cm 0.75cm}, clip]{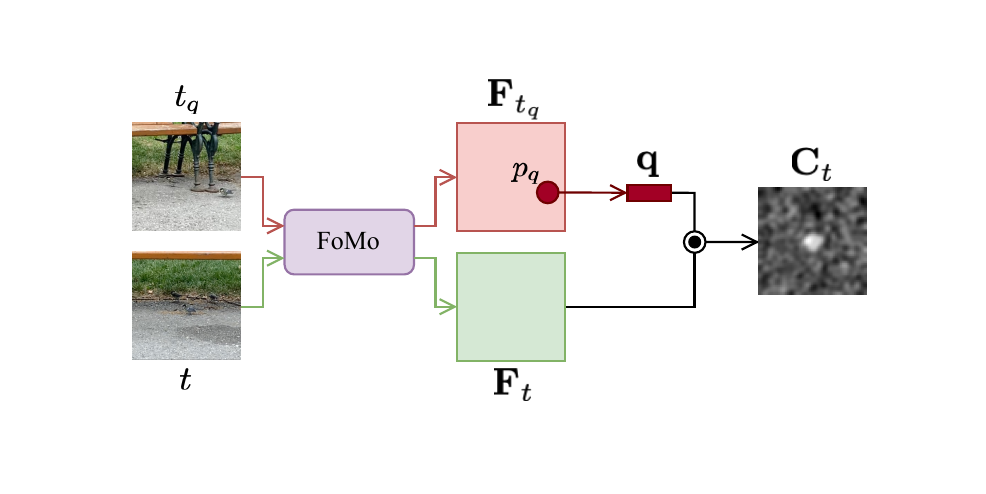}
    \caption{\textbf{Correlation Map.} 
    We compute a dense correlation map $\bC_t$ by measuring cosine similarity between a query feature $\bq$ and all spatial locations in the frame feature map $\bF_t$. The resulting map encodes the similarity distribution for a single point across the frame at time $t$. High similarity regions (highlighted in warmer colors) indicate likely positions for the tracked point. This formulation enables correspondence estimation without explicit supervision and serves as the foundation for both zero-shot retrieval and learning-based decoding.}
    \label{fig:cost_volume}
\end{figure}
In this section, we explain how we leverage foundation models to estimate correspondence. For all settings, we utilize correlation maps to achieve this. Formally, given an encoded video, \ie feature maps, represented by a selected foundation model, $\bF \in \mathbb{R}^{T \times D \times H \times W}$, and a query prompted at frame $t_q$ at location $\bp_q$, we extract the query feature $\mathbf{q} = \bF_{t_q}(\bp_q) \in \mathbb{R}^{D \times 1 \times 1}$ by bilinear interpolation. The correlation map at any time $t$, $\bC_t \in \nR ^{H \times W}$, is then calculated using the cosine similarity between $\bF_t$ and $\mathbf{q}$~(see~\figref{fig:cost_volume}):
\begin{ceqn}
\begin{equation}
    \bC_t = \dfrac{\bq~.~\bF_t}{\| \bq \|~.~\| \bF_t \|} 
\end{equation}
\end{ceqn}

As different models have varying stride values (8 for SD, 14 for the DINO family, and 16 for others), we ensure a fair evaluation by maintaining the same size of the correlation map, \ie the resolution of the final feature map, across different models. This is achieved by resizing the input images so that the feature maps have a consistent $32 \times 32$ resolution. \\

\boldparagraph{Zero-Shot Evaluation} In the zero-shot setting, we directly consider the most similar feature location to the query as the prediction for the corresponding time step, $t$. In other words, the predicted location $\hat{\bp}_t$ is the index of the highest value in $\bC_t$:
\begin{ceqn}
\begin{equation}
    \hat{\bp}_t = \underset{\bp}{\mathrm{argmax}}\, \bC_t(\bp) 
\end{equation}
\end{ceqn}
However, this formulation does not provide information about the visibility of the point and is valid only if the point is visible. Therefore, we evaluate only the visible points. \\

\begin{figure}[t]
    \centering
    \includegraphics[width=.8\linewidth]{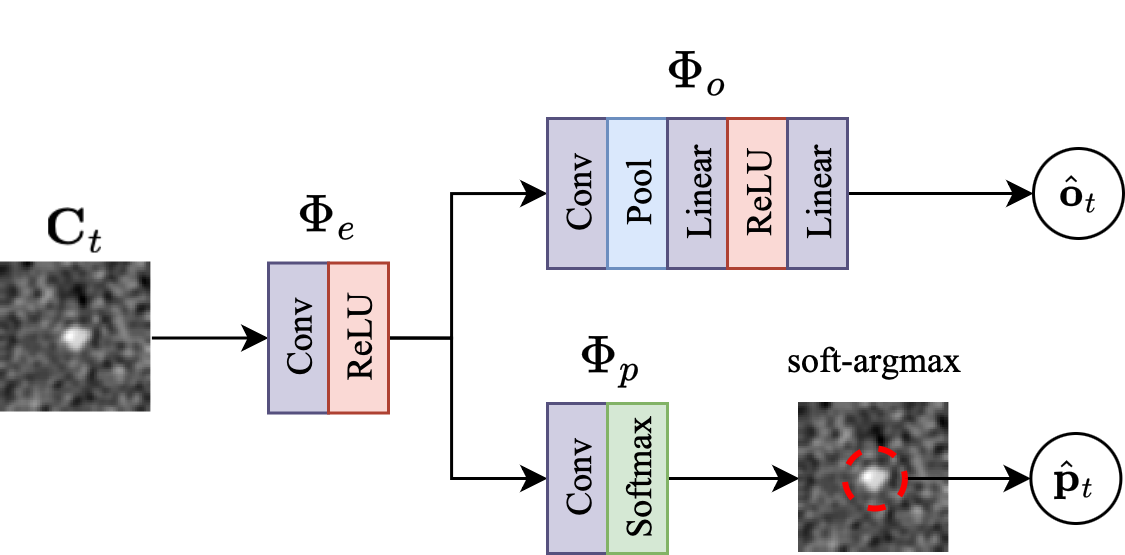}
    \caption{\textbf{Probing the Correlation Map.} 
    We adopt lightweight convolutional heads, inspired by TAPNet~\cite{Doersch2022NeurIPS}, to decode correlation maps into point predictions $\hat{\bp}_t$ and occlusion logits $\hat{\bo}_t$. A shared encoder $\Phi_e$ extracts compact features, which are processed by two specialized heads: one for occlusion classification, and one for localization via soft-argmax.}
    \label{fig:vitap}
\end{figure}

\boldparagraph{Probing}
In TAPNet~\cite{Doersch2022NeurIPS}, correlation maps are computed from backbone features and processed through a small network to decode visibility and point location. The simplicity of this setup makes it ideal for probing whether geometric correspondence is inherently encoded in a frozen feature space.

Inspired by this, we attach low-capacity convolutional branches—comprising only 5.5K parameters—on top of the correlation map $\bC_t$, computed from a frozen foundation model. Formally, the point prediction $\hat{\bp}_t$ and occlusion probability $\hat{\bo}_t$ are computed as:
\begin{ceqn}
\begin{equation}
\begin{aligned}
    \hat{\bp}_t &= \text{soft-argmax}(\Phi_{p} \circ \Phi_{e}(\bC_t)) \\
    \hat{\bo}_t &= \sigma(\Phi_{o} \circ \Phi_{e}(\bC_t))
\end{aligned}
\end{equation}
\end{ceqn}
Here, $\Phi_e$ denotes a shared convolutional layer applied to the correlation map. The outputs are then passed to two task-specific heads: $\Phi_p$ for localizing the target point, and $\Phi_o$ for occlusion classification. The point head uses a convolution followed by a spatial softmax and a soft-argmax to produce a 2D coordinate, while the occlusion head includes a convolution, pooling, and two linear layers with ReLU activation to predict an occlusion logit. We illustrate this architecture in Figure~\ref{fig:vitap}.

The overall design ensures minimal overhead while allowing the model to decode geometric signals, if present, from frozen backbone features. The training objectives include a Huber loss for point localization and a binary cross-entropy loss for occlusion.

\boldparagraph{Adaptation} In the adaptation phase, we go beyond mere probing and update the model to examine whether these models provide a good starting point for learning correspondence in a constrained learning setup. Rather than fine-tuning the entire model, we employ Low Rank Adaptation (LoRA)~\cite{Hu2022ICLR}. Formally, given a pretrained weight $\bW_{\{Q, V\}} \in \nR^{d \times d}$, we use the adapted weight $\bW'$:
\begin{ceqn}
\begin{equation}
    \bW'_{\{Q, V\}}= \bW_{\{Q, V\}} + \bB \bA 
\end{equation}
\end{ceqn}
where $\bA \in \nR^{r \times d}$, $\bB \in \nR^{d \times r}$, $r$ is the rank, and $d$ is the feature dimension. We train only the $\bB$ and $\bA$ matrices while keeping the original $\bW$ matrices frozen, for each attention layer.

This approach maintains the integrity of the pretrained model's information while enabling adaptation with significantly fewer parameters compared to full model fine-tuning. Specifically, we apply LoRA to the query and value projections within the attention layers of the Vision Transformer (ViT), following the methodology in MeLo~\cite{Zhu2023ARXIV}. 

\section{Experiments}

\subsection{Experimental Setup}
\boldparagraph{Datasets} We evaluate the models on the TAP-Vid benchmark, utilizing three datasets: (i) \textbf{TAP-Vid DAVIS} includes 30 real-world videos, each with around 100 frames, featuring intricate motions; (ii) \textbf{TAP-Vid RGB-Stacking} is a synthetic dataset containing 50 videos of robotic manipulation tasks; (iii) \textbf{TAPVid-Kinetics} consists of over 1000 online videos with various actions. For training in probing and adaptation, we use \textbf{TAP-Vid Kubric}, a synthetic simulation dataset of 11K videos, each with a fixed length of 24 frames.

\boldparagraph{Metrics} Consistent with prior work~\cite{Doersch2023ICCV, Doersch2024ARXIV, Karaev2024ECCV}, we evaluate tracking performance using multiple metrics. Occlusion Accuracy (OA) evaluates the correctness of occlusion predictions. $\delta^x_{avg}$ indicates the proportion of visible points tracked within 1, 2, 4, 8, and 16 pixels, averaged across them. Average Jaccard (AJ) combines both occlusion and prediction accuracy. Evaluations are performed in a queried-first mode, where the first visible point for each trajectory serves as the query.

\subsection{Zero-Shot Evaluation}

\begin{table}[t]
    \centering
    \caption{\textbf{Zero-Shot Evaluation.} These results show the zero-shot evaluation results on the TAP-Vid datasets. $\delta^x_{avg}$ is reported.}
    \begin{tabular}{l | c c c | r }
        \toprule
        \textbf{Model} & DAVIS & RGB-St. & Kinetics & \textbf{Avg.}\\
        \midrule
        MAE~\cite{He2022CVPR} & 23.5 & 43.2 & 27.6 & 31.4 \\
        DeIT III~\cite{Touvron2022ECCV} & 24.0 & 23.3 & 22.4 & 23.2 \\
        CLIP~\cite{Radford2021ICML} & 25.4 & 33.8 & 25.0 & 28.1 \\
        SAM~\cite{Kirillov2023ICCV} & 29.5 & \underline{44.7} & 31.4 & 35.2\\
        SD~\cite{Rombach2022CVPR} & 33.9 & \textbf{46.2} & \textbf{37.2} & \textbf{39.1} \\
        DINO~\cite{Caron2021ICCV} & 34.5 & 39.3 & 34.4 & 36.1 \\
        DINOv2-Reg~\cite{Darcet2023ARXIV} & \underline{37.4} & 35.9 & 33.6 & 35.6 \\
        DINOv2~\cite{Oquab2024TMLR}  & \textbf{38.0} & 37.8 & \underline{34.5} & \underline{36.8}\\
        \midrule
        \color{gray} TAPNet~\cite{Doersch2022NeurIPS} & \color{gray} 48.6 & \color{gray} 68.1 & \color{gray} 54.4 & \color{gray} 57.0 \\
        \bottomrule
    \end{tabular}
    \label{tab:eval_fomo:zero_shot}
\end{table}

\begin{figure}[!h]
    \centering
    \begin{minipage}{0.49\textwidth}
        \subfloat{\includegraphics[width=.99\linewidth, trim={0cm 1.5cm 0cm 2cm}, clip]{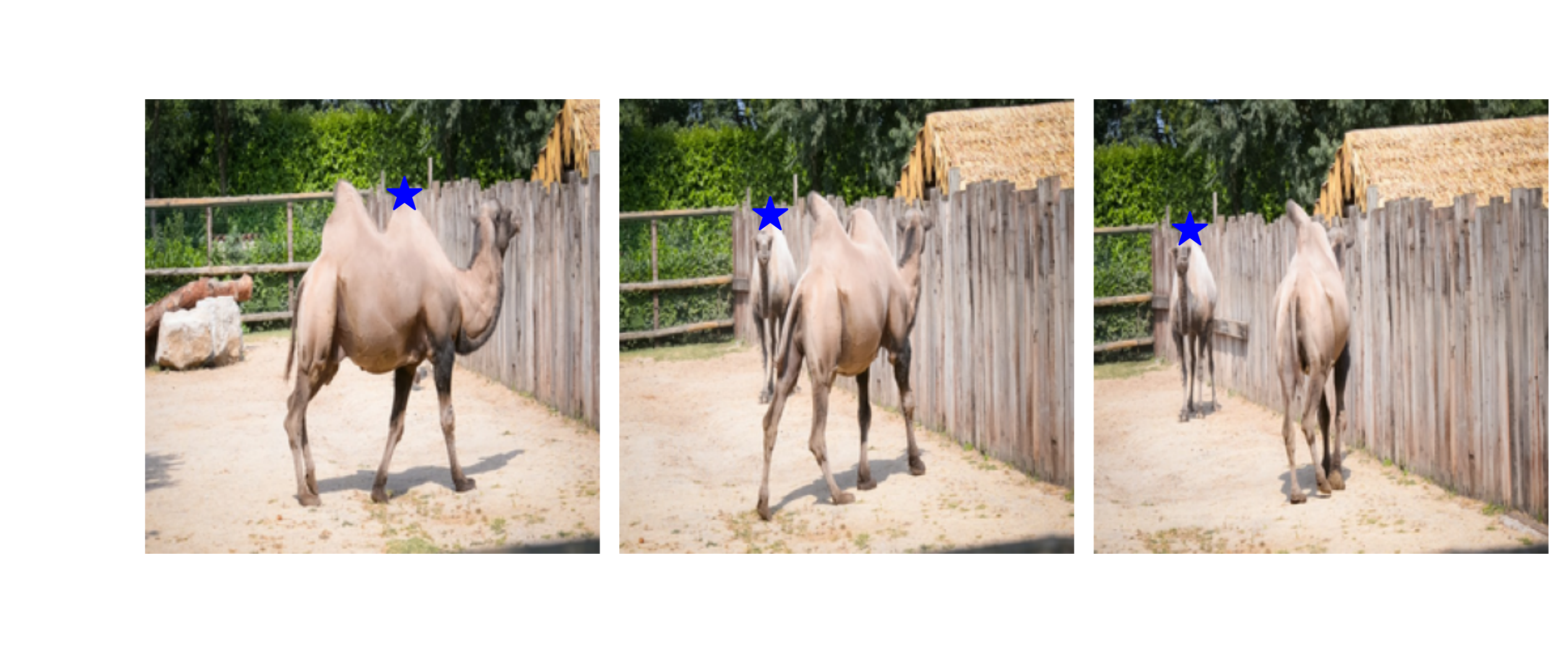}}\vspace{-0.1cm}
        \subfloat{\includegraphics[width=.99\linewidth, trim={0cm 1.5cm 0cm 2cm}, clip]{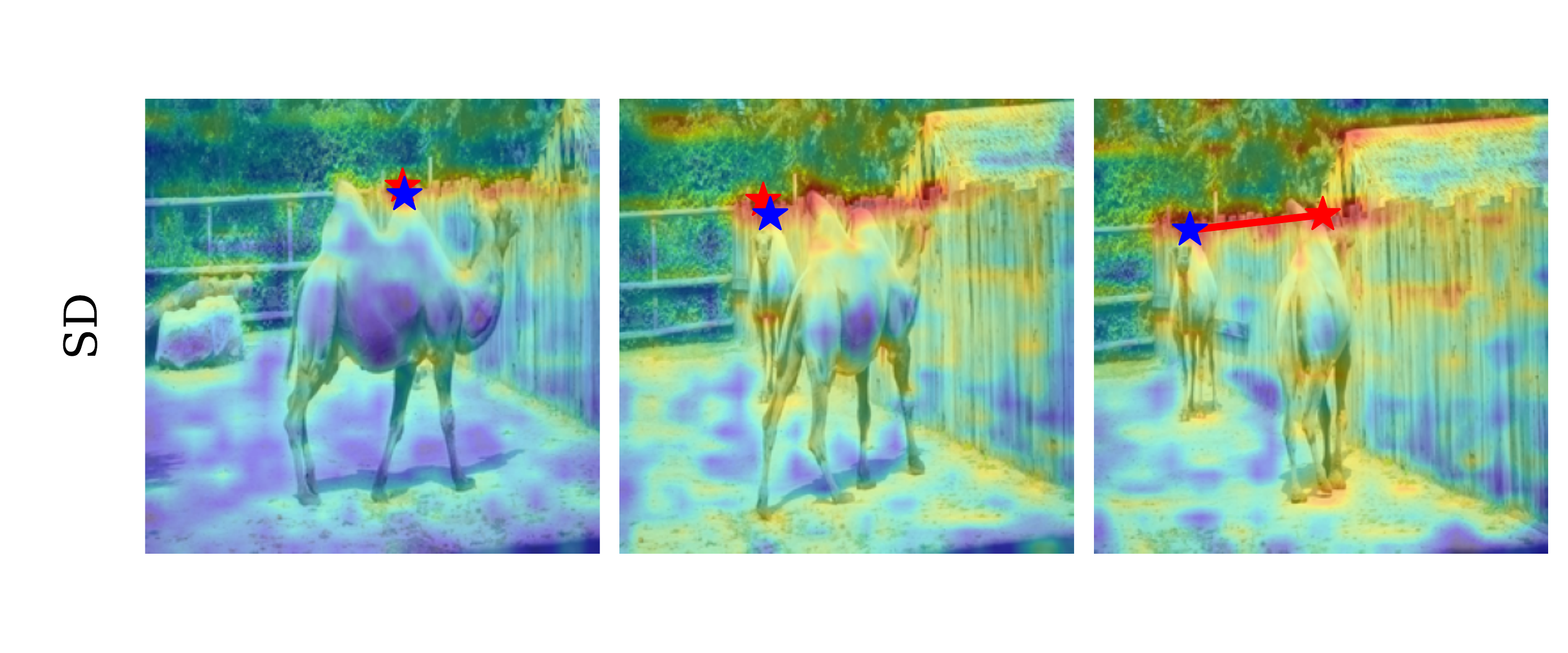}}\vspace{-0.1cm}
        \subfloat{\includegraphics[width=.99\linewidth, trim={0cm 1.5cm 0cm 2cm}, clip]{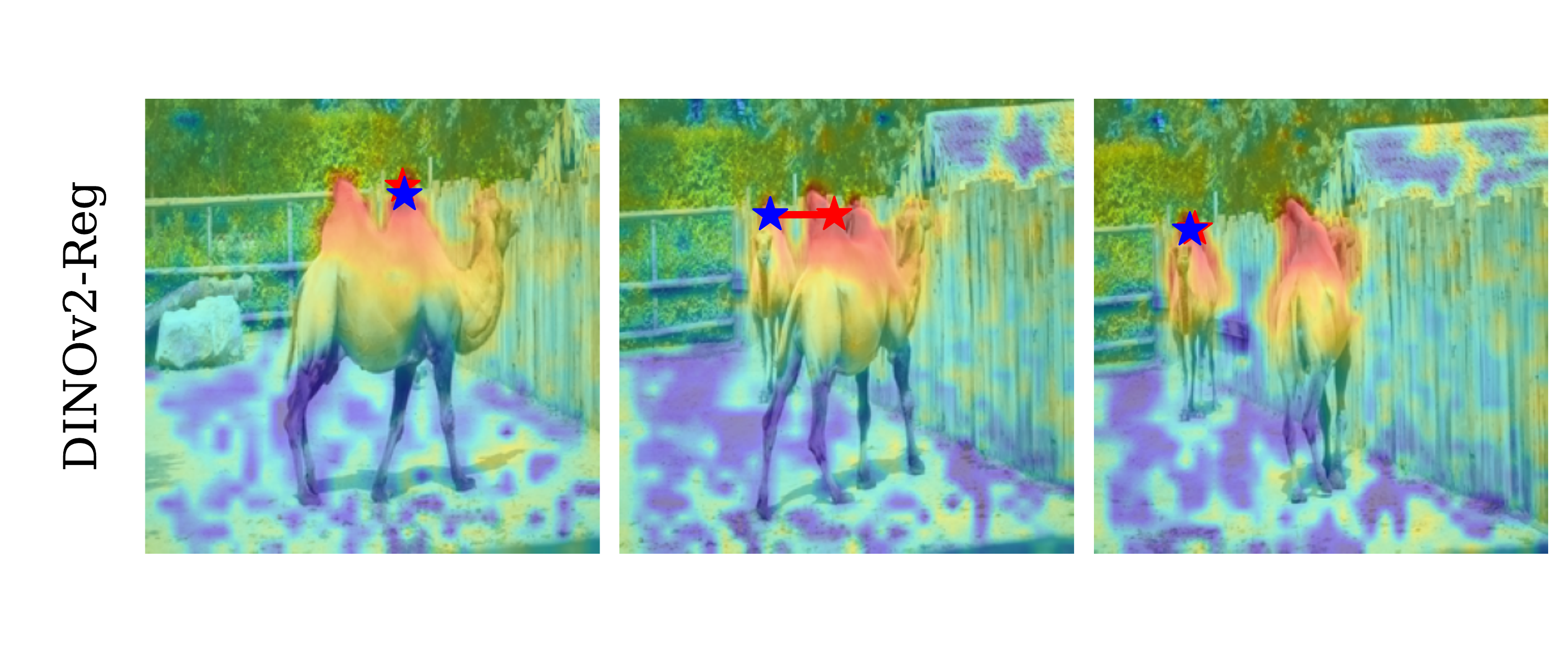}}\vspace{-0.1cm}
        \subfloat{\includegraphics[width=.99\linewidth, trim={0cm 1.5cm 0cm 2cm}, clip]{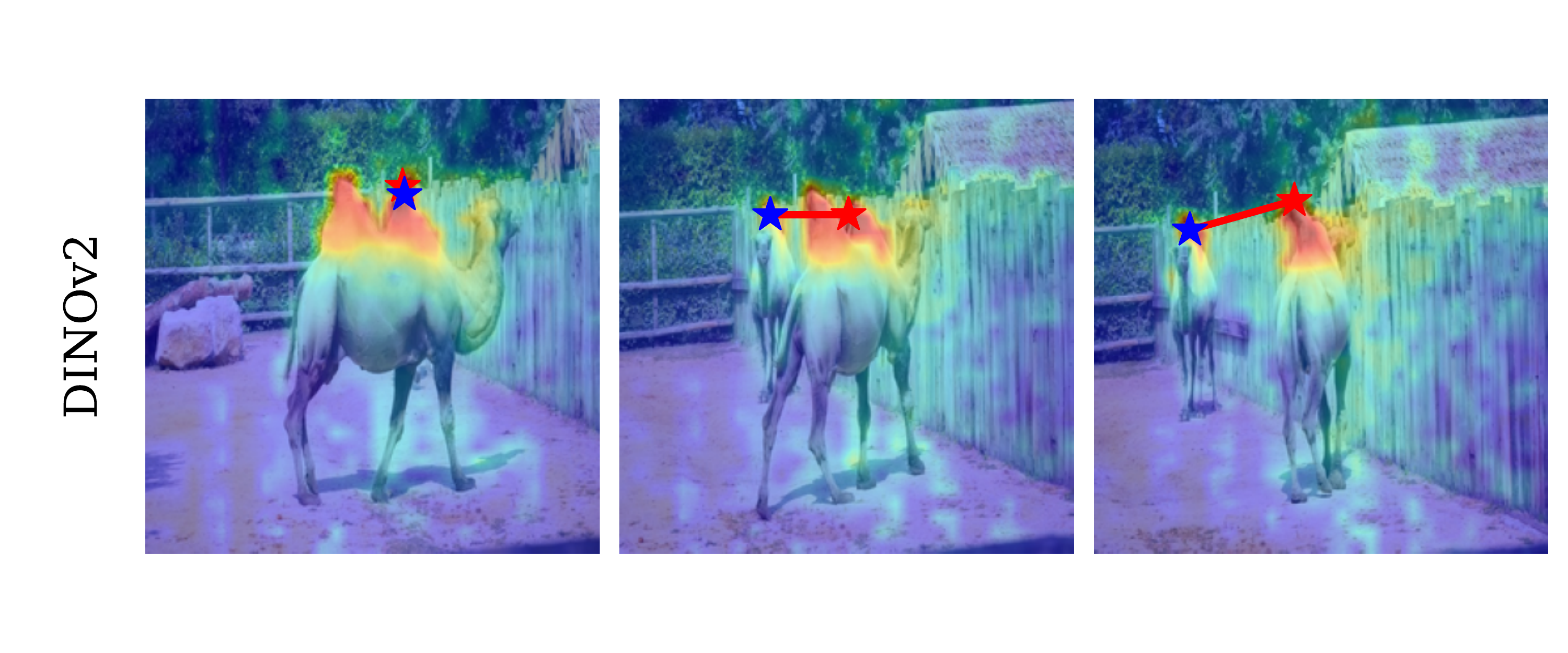}}  \vspace{-0.1cm}
        \subfloat{\includegraphics[width=.99\linewidth, trim={0cm 1.5cm 0cm 2cm}, clip]{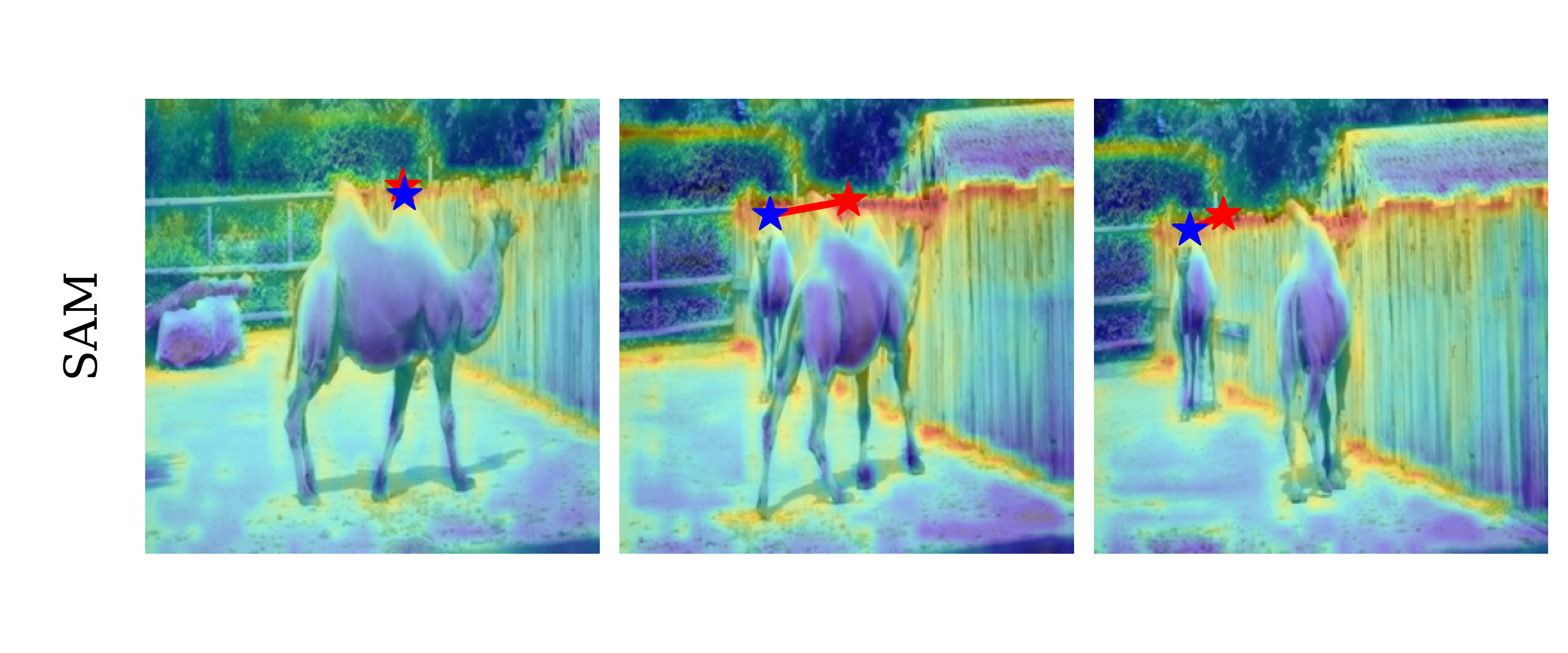}}  
    \end{minipage} \hfill
    \begin{minipage}{0.49\textwidth}
        \subfloat{\includegraphics[width=.99\linewidth, trim={0cm 1.5cm 0cm 2cm}, clip]{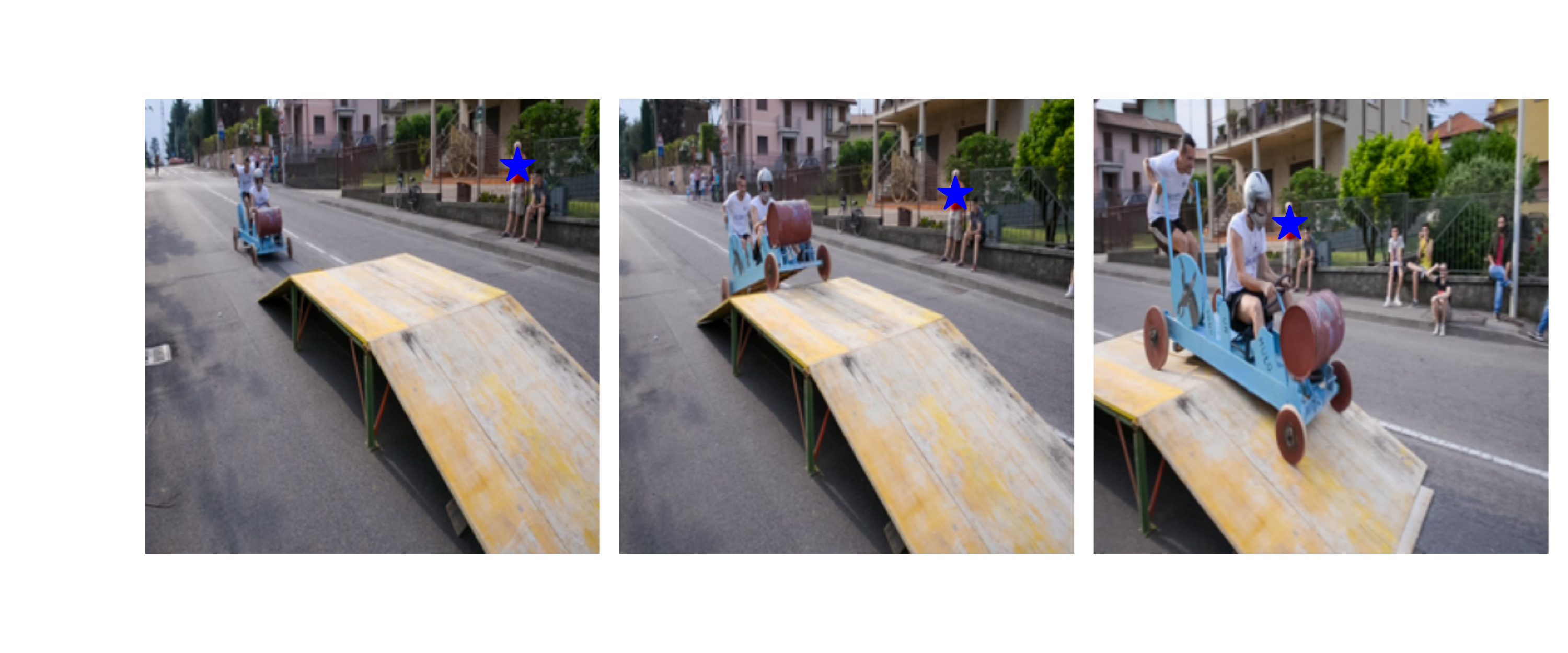}}\vspace{-0.1cm}
        \subfloat{\includegraphics[width=.99\linewidth, trim={0cm 1.5cm 0cm 2cm}, clip]{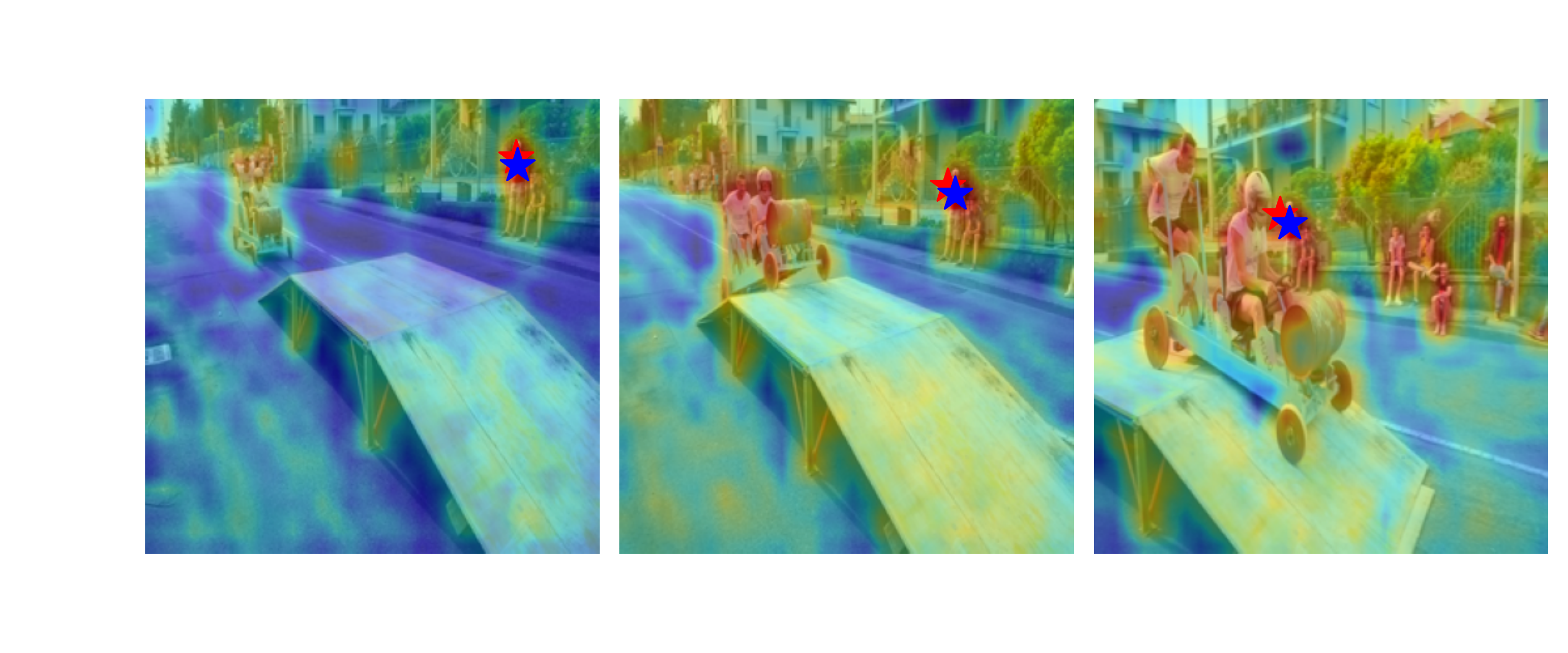}}\vspace{-0.1cm}
        \subfloat{\includegraphics[width=.99\linewidth, trim={0cm 1.5cm 0cm 2cm}, clip]{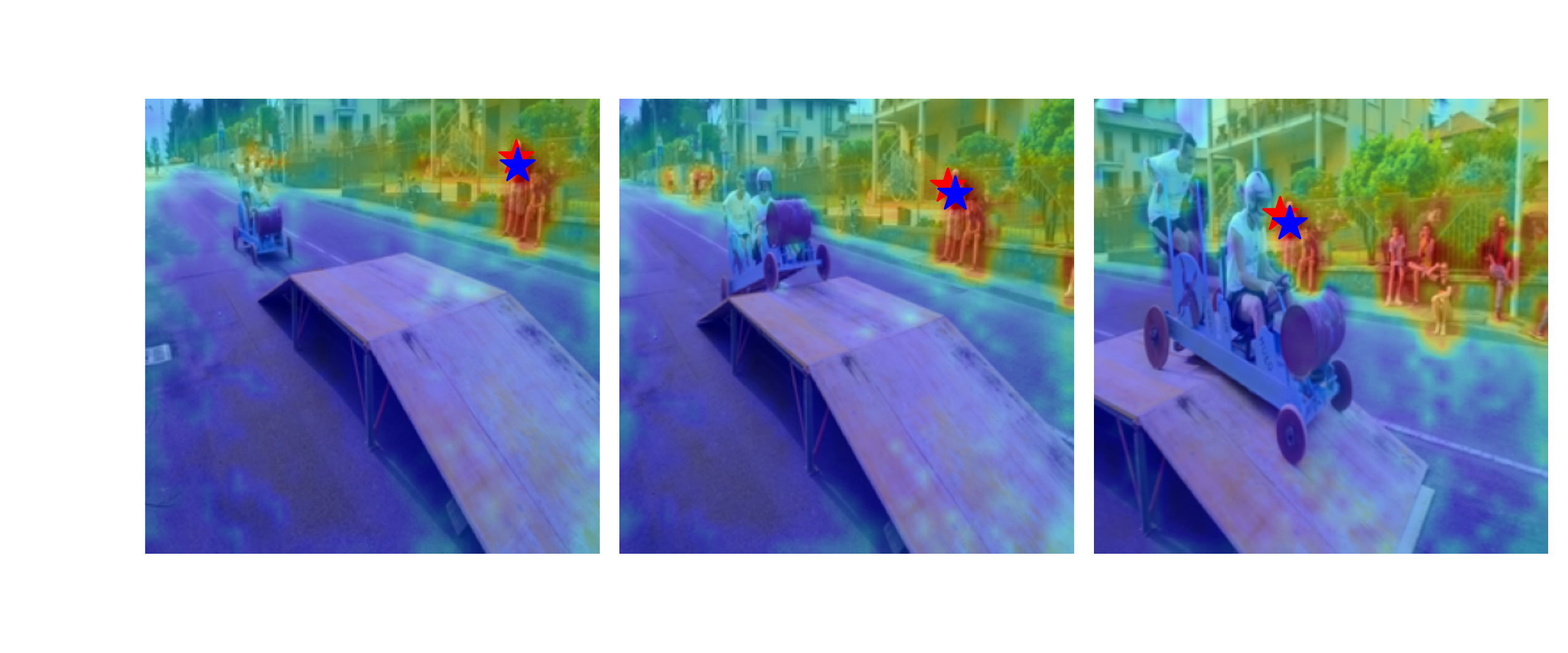}}\vspace{-0.1cm}
        \subfloat{\includegraphics[width=.99\linewidth, trim={0cm 1.5cm 0cm 2cm}, clip]{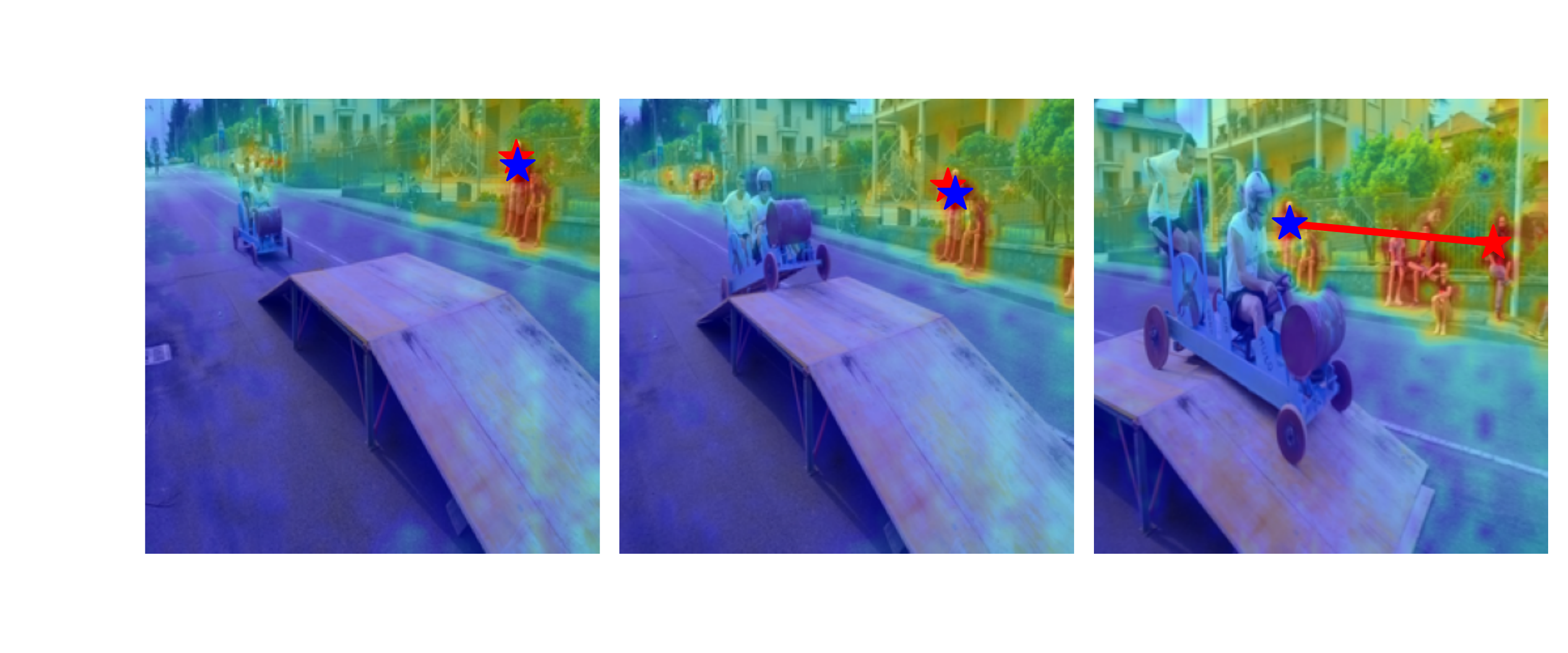}}  \vspace{-0.1cm}
        \subfloat{\includegraphics[width=.99\linewidth, trim={0cm 1.5cm 0cm 2cm}, clip]{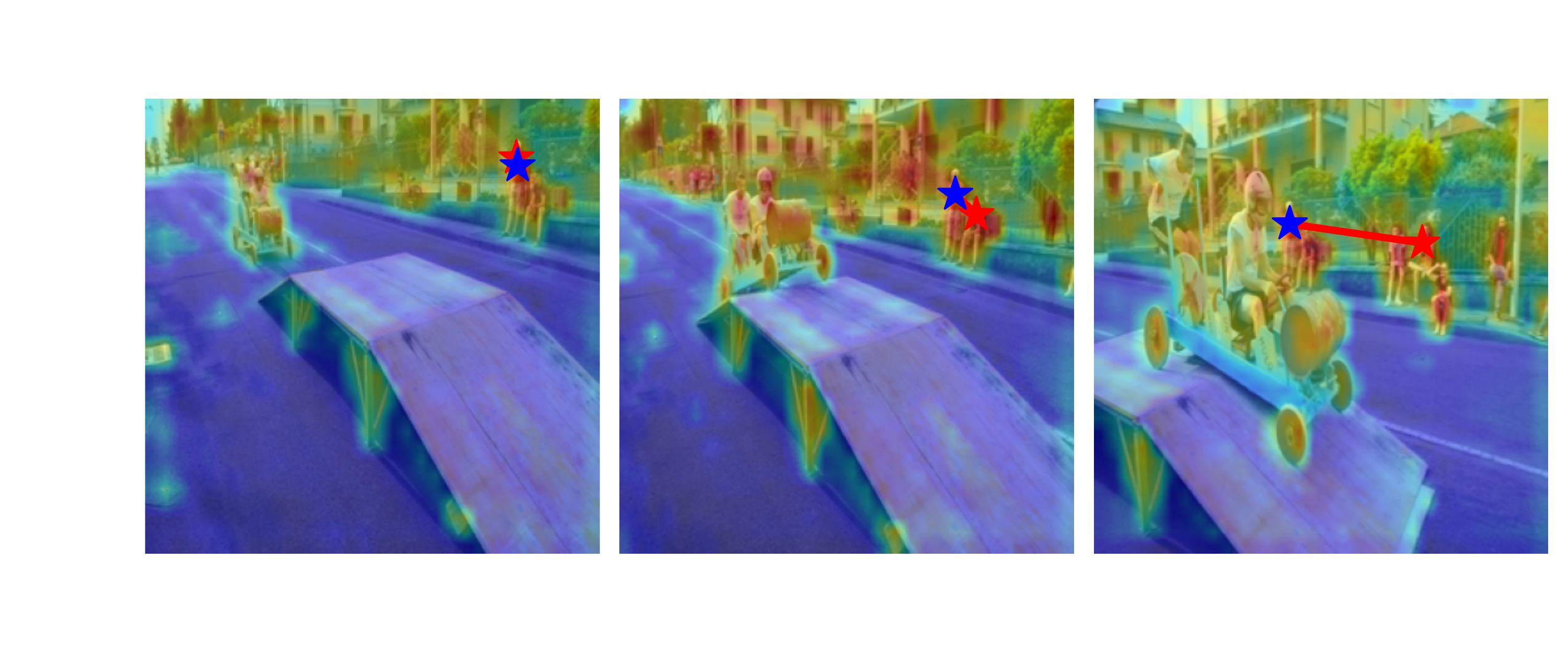}}
    \end{minipage}%

    \caption{\textbf{Zero-Shot Tracking on TAP-Vid DAVIS.} We show correlation maps $\bC_t$ for query points from two videos using Stable Diffusion~\cite{Rombach2022CVPR}, DINOv2~\cite{Oquab2024TMLR}, DINOv2-Reg~\cite{Darcet2023ARXIV}, and SAM~\cite{Kirillov2023ICCV}. We compute correlation maps between a sampled query feature and frame features at later timesteps. Predictions ({\color{red} red stars}) and ground truth ({\color{blue} blue stars}) are connected by red lines to indicate tracking error. Warmer colors in the correlation maps indicate stronger feature similarity.}
    \label{fig:eval_fomo:qualitative}
\end{figure}

\boldparagraph{Different Models} The results of different foundation models in the zero-shot setting are presented in \tabref{tab:eval_fomo:zero_shot}. We also report the performance of a supervised model, TAPNet~\cite{Doersch2022NeurIPS}, as it is the only supervised model with a final feature map resolution of $32 \times 32$. The DINO family demonstrates superior performance on DAVIS. Conversely, SD and SAM outperform the DINO models on RGB-Stacking. One possible reason for this is that RGB-Stacking is a synthetic dataset, giving SD an advantage due to its more diverse training set, which includes both real-world and synthetic data. Additionally, SD performs the best on Kinetics. On average, SD achieves the highest performance, followed by DINOv2. We visualize correlation maps of SD, DINOv2, DINOv2-Reg, and SAM in \figref{fig:eval_fomo:qualitative}. SD appears to have greater geometric sensitivity, while DINOv2 demonstrates higher semantic capability. Overall, the superior performance of these two models aligns with prior work~\cite{El2024CVPR, Zhan2023ARXIV}, which shows that DINOv2 and SD have a better 3D understanding compared to other models. 

\boldparagraph{Effect of Architectural Variations} 
We conducted an ablation study on: (i) architectural differences to understand how the backbone capacity of the same pre-training method affects correspondence awareness, and (ii) input resolution to examine the effect of resolution, a key factor in correspondence~\cite{Doersch2023ICCV}, in \tabref{tab:eval_fomo:zero_shot_arch} using DINOv2. The results indicate that using larger architectures consistently improves performance, with an increase from 37.1 (ViT-S) to 40.0 (ViT-G). Additionally, the resolution of the final feature map has a significant impact. Higher resolutions lead to a performance gain of 10.5 ($64 \times 64$), while lower resolutions result in a reduction of 13.1 ($16 \times 16$).

\begin{table}[h]
    \centering
    \caption{\textbf{Effect of Architecture for Zero-Shot Evaluation.} This table shows the zero-shot evaluation results for different architectures of DINOv2 by varying the resolution on TAP-Vid DAVIS.}
    \begin{tabular}{l c | c}
        \toprule
        \textbf{Architecture} & \textbf{Final Resolution} & $\delta^x_{avg}$  \\
        \midrule
        ViT-\textbf{S}/14 & \multirow{4}{*}{32 $\times$ 32} & 37.1 {\color{red} \scriptsize (-0.9)}  \\
        ViT-\textbf{B}/14 &  & 38.0 \\
        ViT-\textbf{L}/14 &  & 39.1 {\color{ForestGreen} \scriptsize (+1.1)} \\
        ViT-\textbf{G}/14 &  & 40.0 {\color{ForestGreen} \scriptsize (+2.0)}\\
        \midrule
        \multirow{3}{*}{ViT-\textbf{B}/14} & 16 $\times$ 16 & 24.9 {\color{red} \scriptsize (-13.1)}\\
         & 48 $\times$ 48 & 45.1 {\color{ForestGreen} \scriptsize (+7.1)} \\
         & 64 $\times$ 64 & 48.5 {\color{ForestGreen} \scriptsize (+10.5)}\\
        
        \bottomrule
    \end{tabular}
    \label{tab:eval_fomo:zero_shot_arch}
\end{table}

\subsection{Probing and Adaptation}
\quad We chose DINOv2 for probing and adaptation due to its top-2 performance in zero-shot evaluation and its resource efficiency compared to SD. For computational efficiency, we selected the ViT-S/14 architecture. For both settings, we used AdamW~\cite{Loshchilov2019ICLR}, a batch size of 16 (1/4 of TAPNet), a learning rate of $1 \times 10^{-3}$, linear warm-up, and cosine decay.  We trained for 20 epochs (\mytexttilde 1/4 iterations of TAPNet) for probing and 40 epochs (\mytexttilde 1/2 iterations of TAPNet) for adaptation. The applied weight decays were $1 \times 10^{-3}$ for probing and $1 \times 10^{-5}$ for adaptation.

We compared probing and adaptation results with different ranks, as shown in~\tabref{tab:eval_fomo:probing}. By only probing correlation maps, DINOv2 can surpass TAPNet in OA. Moreover, adaptation of any rank performs better than TAPNet across all metrics. Surpassing supervised models even in a significantly constrained training setup, where the number of learnable parameters is substantially lower (2.5\% for rank 16), demonstrates that DINOv2 could serve as a strong initialization for correspondence learning.

\begin{table}[t]
    \centering
    \caption{\textbf{Probing and Adapting DINOv2.} This table shows different setups for DINOv2 ViT-S/14 on the TAP-Vid DAVIS, including the number of learnable parameters. The setups include zero-shot, probing, and adaptation with various LoRA ranks.}
    \begin{tabular}{l c c c | c c c}
        \toprule
        \textbf{Model} & \textbf{Setup} & \textbf{Rank} & \textbf{\#L.P.} & \text{AJ} & $\delta^x_{avg}$   & \text{OA} \\
        \midrule
        \multirow{5}{*}{DINOv2} & \text{Zero-shot} & - & 0 & - & 37.1 & - \\
         & \text{Probing} & - & 5.5K & 27.1 & 42.3 & 79.4 \\
         & \text{Adaptation} & 16 & 0.3M & 33.4 & 49.0 & 80.1 \\
         & \text{Adaptation} & 32 & 0.6M & 33.9 & 49.7 & \textbf{80.4} \\
         & \text{Adaptation} & 64 & 1.2M & \textbf{35.0} & \textbf{51.3} & 80.2 \\
        \midrule
        TAPNet & \text{Supervised} & - & 12.0M &  33.0 &  48.6 &  78.8 \\
        \bottomrule
    \end{tabular}
    \label{tab:eval_fomo:probing}
\end{table}

\subsection{Discussion}
We explored the geometric awareness of vision foundation models for long-term point tracking under zero-shot settings and showed that Stable Diffusion and DINOv2 have better geometric understanding than other models. Moreover, our experiments demonstrate that DINOv2 can surpass the performance of supervised models despite a significantly lighter training setup, indicating that these models inherently possess the notion of correspondence.

\chapter{Track-On: Transformer-based Online Point Tracking with Memory}
\label{chapter:track_on}
In Chapter~\ref{chapter:eval_fomo}, we showed that visual foundation models encode strong geometric priors and can be repurposed for point tracking via simple correlation-based formulations. However, despite their success in zero-shot and lightly supervised settings, these models lack temporal reasoning and operate without memory, limiting their performance in online tracking scenarios. More broadly, existing point tracking methods, such as TAPIR~\cite{Doersch2023ICCV} and CoTracker~\cite{Karaev2024ECCV}, are designed for offline use. They process multiple frames jointly, often relying on future context, making them unsuitable for streaming video where predictions must be made sequentially.

While FoMos can be integrated into these models to improve accuracy, this does not resolve the fundamental challenge: tracking in an online setting. Even with strong features, existing models struggle to maintain consistency over time without access to future frames or full spatiotemporal attention. This motivates the need for an architecture explicitly designed for online long-term tracking.

In this chapter, we introduce~\textbf{Track-On}, a transformer-based model that tracks points causally, frame by frame. Each point is represented as a query in the decoder, and updated through two specialized memory modules. The spatial memory stores localized features around past predictions to reduce drift, while the context memory aggregates point-wise embeddings over time to maintain continuity. This design enables accurate and efficient long-term tracking under strict online constraints.

\section{Methodology}

\subsection{Problem Scenario}
Given an RGB video of $T$ frames, $\cV = \bigl\{\bI_1,~ \bI_2,~ \dots,~ \bI_T \bigr\} \in \nR^{T \times H \times W \times 3}$, and a set of $N$ predefined queries, $\cQ=  \bigl\{ (t^{1}, \bp^{1}),~(t^{2}, \bp^{2}),~ \dots,~ ~(t^{N}, \bp^{N}) \bigr\}\in \nR^{N \times 3}$, where each query point is specified by the start time and pixel's spatial location, our goal is to predict the correspondences $\hat{\bp}_t \in \nR^{N \times 2}$ and visibility $\hat{\bv}_t \in \{0, 1\}^{N}$ for all query points in an online manner, \ie using only frames up to the current target frame $t$. To address this problem, we propose a transformer-based point tracking model, that tracks points \textbf{frame-by-frame}, 
with dynamic memories $\bM$ to propagate temporal information along the video sequence:
\begin{ceqn}
\begin{equation}
\hat{\bp}_t,~ \hat{\bv}_t,~ \bM_t = \Phi \left(\bI_t,~ \cQ,~ \bM_{t-1};~ \Theta  \right)
\end{equation}
\end{ceqn}
In the following sections, we start by describing the basic transformer architecture for point tracking in \secref{sec:track_on:vanilla_model}, then introduce the two memory modules and their update mechanisms in \secref{sec:track_on:full_model}.

\begin{figure}[t]
    \centering
    \includegraphics[width=1\linewidth]{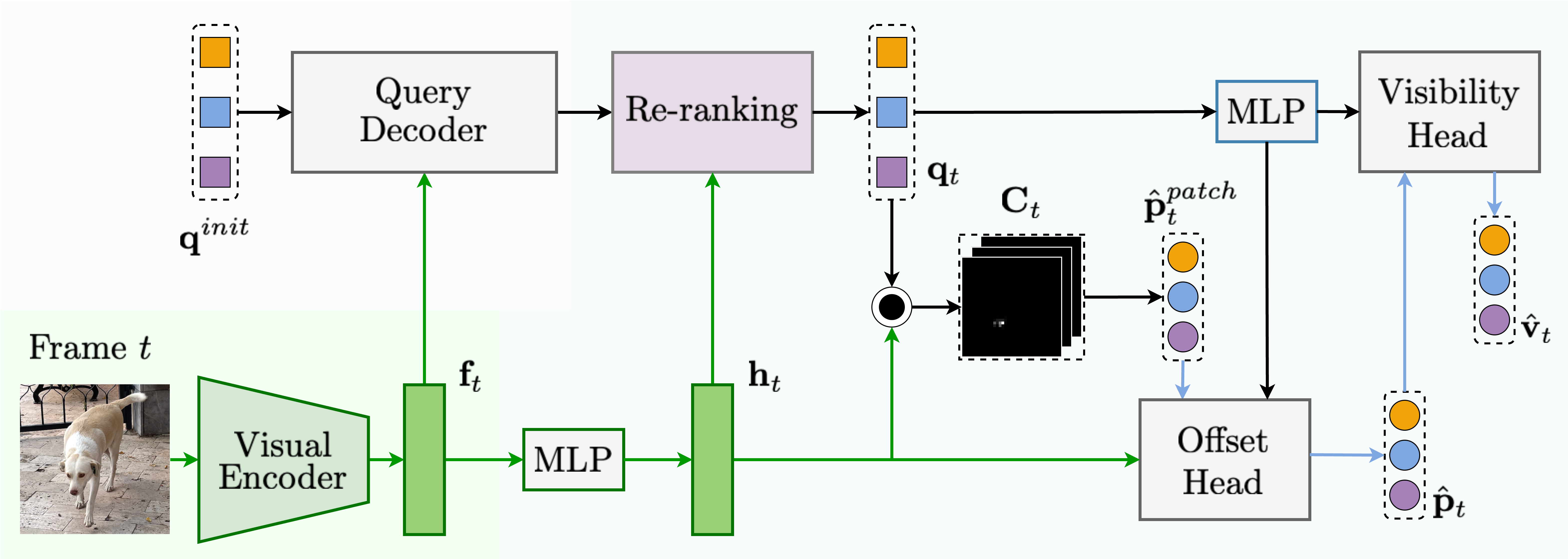}
    \caption{\textbf{Overview.} We introduce Track-On, a simple transformer-based method for online, frame-by-frame point tracking. The process involves three steps: (i) \textbf{Visual Encoder}, which extracts features from the given frame; (ii) \textbf{Query Decoder}, which decodes interest point queries using the frame’s features; (iii) \textbf{Point Prediction}~(highlighted in light blue),
    where correspondences are estimated in a coarse-to-fine manner, 
    first through patch classification based on similarity, then
    followed by refinement through offset prediction from a few most likely patches. 
    Note that the squares refer to point queries, while the circles represent predictions, either as point coordinates or visibility.}
    \label{fig:track_on:main_fig}
\end{figure}

\subsection{Track-On: Point Tracking with a Transformer}
\label{sec:track_on:vanilla_model}
Our model is based on transformer, consisting of three components,
as illustrated in \figref{fig:track_on:main_fig}:  \textbf{Visual Encoder} is tasked to extract visual features of the video frame, and initialize the query points; \textbf{Query Decoder} enables the queried points to attend the target frame to update their features; and \textbf{Point Prediction}, to predict the positions of corresponding queried points in a coarse-to-fine manner.

\subsubsection{Visual Encoder}
\label{sec:track_on:vis_encoder}

We adopt a Vision Transformer (ViT) as our visual backbone, specifically DINOv2~\cite{Oquab2024TMLR}, due to its strong viewpoint invariance and adaptation flexibility, as demonstrated in Chapter~\ref{chapter:eval_fomo}. However, standard ViTs process image patches with relatively large stride, which limits spatial resolution and poses challenges for dense prediction tasks. To address this, we employ ViT-Adapter~\cite{Chen2023ICLR} to obtain higher-resolution feature maps. Finally, we add learnable spatial positional embeddings $\gamma^s$ to the per-frame features:
\begin{ceqn}
\begin{equation}
\bff_t = \Phi_{\text{vis-enc}} \left(\bI_t\right) + \gamma^s  \in \nR^{\frac{H}{S} \times \frac{W}{S} \times D}
\end{equation}
\end{ceqn}
where $D$ denotes the feature dimension, and $S$ refers to the stride. 
We use a single-scale feature map from ViT-Adapter for memory efficiency, specifically with a stride of $S = 4$.

\boldparagraph{Query Initialization} 
To initialize the query features~($\bq^{init}$), 
we apply bilinear sampling to the feature map at the query position~$\left(\bp^{i}\right)$:
\begin{ceqn}
\begin{equation}
\bq^{init} = \bigl\{\text{sample}(\bff_{t^{i}},~ \bp^{i})\bigr\}_{i = 1}^{N} \in \nR^{N \times D}
\end{equation}
\end{ceqn}
In practice, we initialize the query based on the features of the start frame $t^{i}$ for $i$-th query, assuming they can start from different time point, and propagate them to the subsequent frames.

\subsubsection{Query Decoder}
\label{sec:track_on:query_decoder}

After extracting the visual features for the frame and query points, 
we adopt a variant of transformer decoder~\cite{Vaswani2017NeurIPS}, 
with 3 blocks, \ie cross-attention followed by a self-attention, with an additional feed forward layer between attentions: 
\begin{ceqn}
\begin{equation}
\bq^{dec}_t = \Phi_{\text{q-dec}} \left( \bq^{init},~ \bff_{t} \right) \in \nR^{N \times D}
\end{equation}
\end{ceqn}
The points of interest are treated as queries, which update their features by iteratively attending to visual features of the current frame with cross attention. These updated queries are then used to search for the best match within the current frame, as explained in the following section.

\subsubsection{Point Prediction}
\label{sec:track_on:point_prediction}

Unlike previous work that regresses the exact location of the points, we formulate the tracking as a matching problem to one of the patches, that provides a coarse estimate of the correspondence. 
For exact correspondence with higher precision, we further predict offsets to the patch center. 
Additionally, we also infer the visibility $\hat{\bv}_t \in [0, 1]^{N}$ and uncertainty $\hat{\bu}_t \in [0, 1]^{N}$ for the points of interest. 

\boldparagraph{Patch Classification} 
We first pass the visual features into 4-layer MLPs, and downsample the resulting features into multiple scales, \ie, $\bff_t \rightarrow \bh_t \in \nR^{\frac{H}{S} \times \frac{W}{S} \times D} \rightarrow \bh_t^l \in \nR^{\frac{H}{2^{l}.S} \times \frac{W}{2^{l}.S} \times D}$. 
We compute the cosine similarity between the decoded queries and patch embeddings in four scales, 
and the similarity map $\bC^{{dec}}_t$ is obtained as the weighted average of multi-scale similarity maps with learned coefficients. We then apply a temperature to scale the similarity map and take softmax spatially over the patches within the current frame. The resulting $\bC^{{dec}}_t$ provides a measure of similarity for each query across the patches in the frame.

We train the model with a classification objective, where the ground-truth class is the patch with the point of interest in it. In other words, we perform a $P$-class classification, $P$ is the total number of patches in the frame.

\begin{figure}[t]
    \centering
    \includegraphics[width=0.65\linewidth]{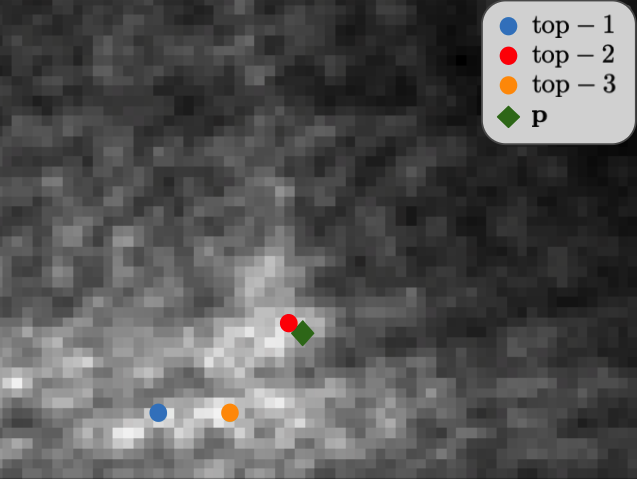}
    \caption{\textbf{Top-$k$ Points.} In certain cases, a patch with high similarity, though not the most similar, is closer to the ground-truth patch. The top-$3$ patch centers, ranked by similarity, are marked with dots, while the ground-truth is represented by a {\color{Green} diamond}.}
    \label{fig:track_on:top_k}
\end{figure}

\boldparagraph{Re-ranking}
We observed that the true target patch might not always have the highest similarity on $\bC^{{dec}}_t$, however, it is usually among the top-k patches. 
For example, in \figref{fig:track_on:top_k}, the patch with the second-highest similarity (top-2) is closer to the true correspondence than the most similar patch (top-1). To rectify such cases, we introduce a re-ranking module $\Phi_\text{re-rank}$:
\begin{ceqn}
\begin{equation} \label{eq:c_t}
     \bq_t = \Phi_\text{re-rank} \left( \bq^{dec}_t,~ \bh_{t},~ \bC^{dec}_t\right) \in \nR^{N \times D}
\end{equation}
\end{ceqn}
where $\bq_t$ denotes the refined queries after ranking. 

\begin{figure}[t]
    \centering
    \includegraphics[width=0.8\linewidth]{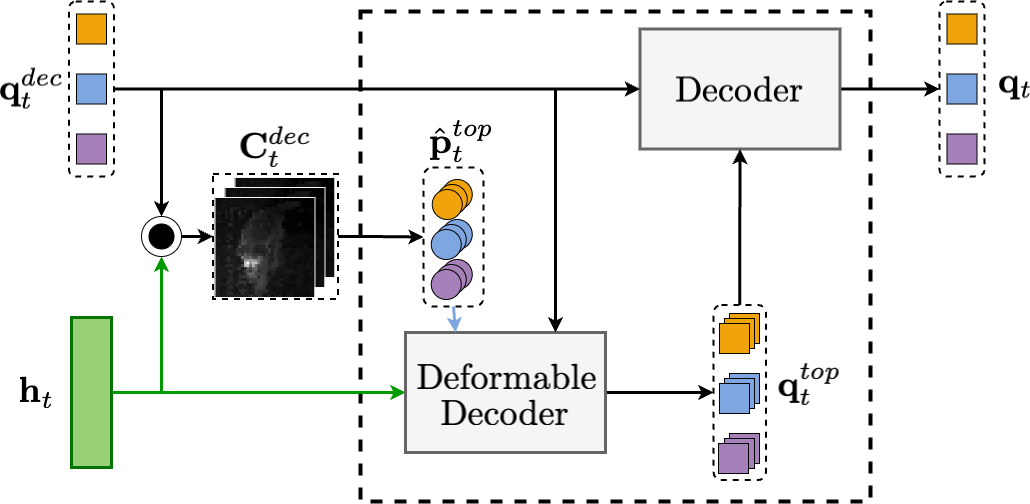}
        \caption{\textbf{Re-ranking Module.}
        The features around the top-$k$ points ($\hat{\bp}_t^{{top}}$) with the highest similarity are decoded using deformable attention to extract the corresponding top-$k$ features ($\bq_t^{{top}}$). These features are then fused with the decoded query $\bq_t^{{dec}}$ using a transformer decoder.}
        \label{fig:track_on:ranking}
\end{figure}

In the re-ranking module (\figref{fig:track_on:ranking}), we identify the top-$k$ patches with the highest similarities and retrieve their corresponding features with a deformable attention decoder. 
Then, we integrate them into the original query features via a transformer decoder to produce refined queries. Using these refined queries, we calculate the final similarity map $\bC_t$ and apply a classification loss. 
Finally, we select the center of the patch with the highest similarity ($\hat{\bp}^{patch}_t \in \nR^{N \times 2}$) as our coarse prediction. Additionally, we compute an uncertainty score for each top-$k$ location, \ie $\hat{\bu}^{top}_t \in \mathbb{R}^{N \times k}$, by processing their corresponding features with a linear layer.

\begin{figure}[t]
    \centering
    \includegraphics[width=0.9\linewidth]{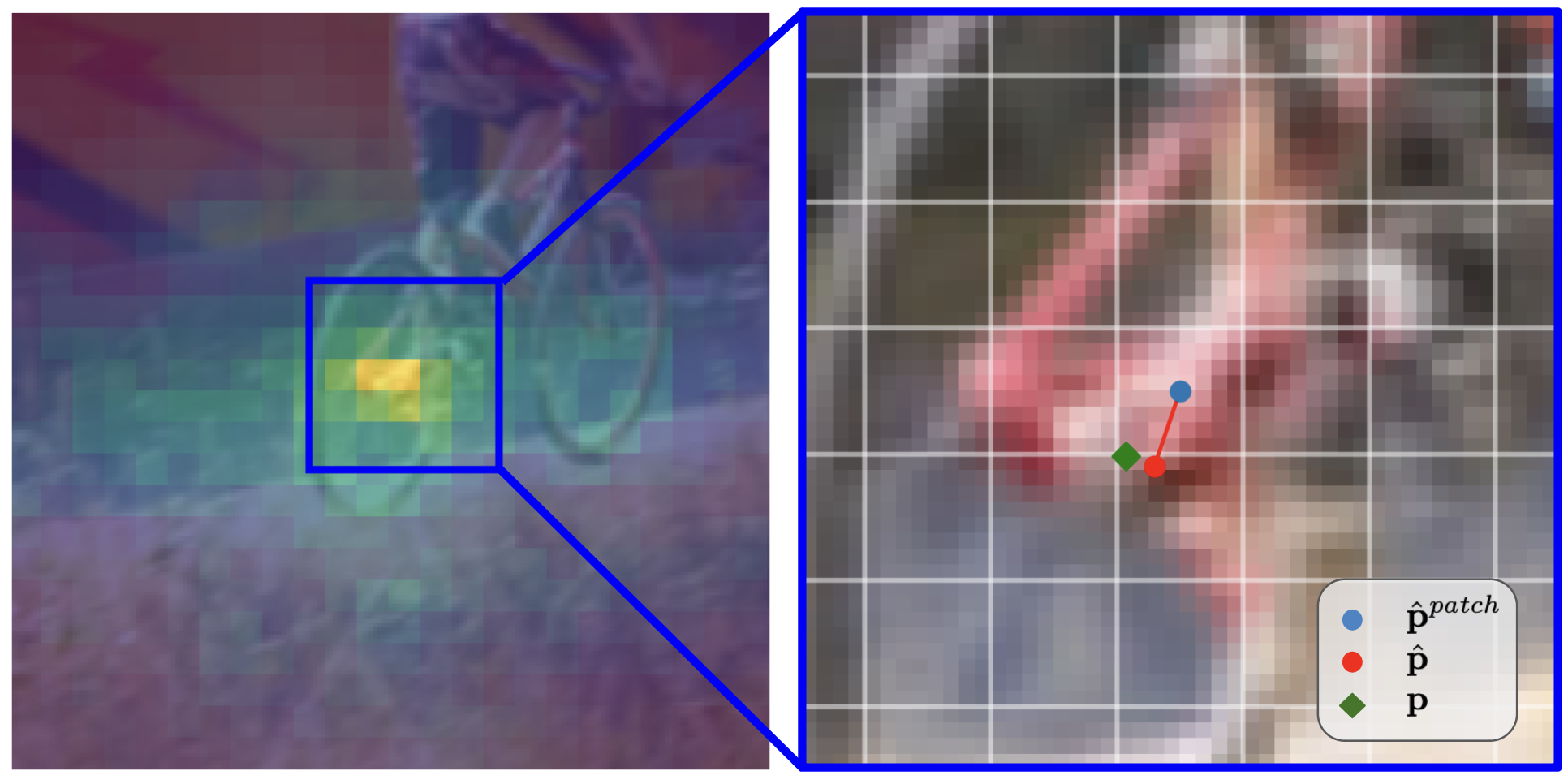}
        \caption{\textbf{Offset Head}. Starting with a rough estimation from patch classification (\textbf{left}), where lighter colors indicate higher correlation, we refine the prediction using the offset head (\textbf{right}). The selected patch center and the final prediction are marked by a {\color{RoyalBlue} blue dot} and a {\color{red} red dot}, respectively, with the ground-truth represented by a {\color{Green} diamond}.}
        \label{fig:track_on:offset}
\end{figure}

\boldparagraph{Offset Prediction}
For the exact correspondence~($\hat{\bp}_t \in \nR^{N \times 2}$), we further predict an offset $\hat{\bo}_t \in \nR^{N \times 2}$ to the patch center by incorporating features from the local region around the inferred patch, as shown in~\figref{fig:track_on:offset}:
\begin{ceqn}
\begin{equation}
\label{eq:patch_refinement}
\hat{\bo}_t = \Phi_{\text{off}}(\bq_t,~\bh_t,~\hat{\bp}^{patch}_t), \quad \quad \quad %
\hat{\bp}_t = \hat{\bp}^{patch}_t + \hat{\bo}_t %
\end{equation}
\end{ceqn}
Here, $\Phi_{\text{off}}$ is a deformable transformer decoder~\cite{Zhu2021ICLR} block with 3 layers, excluding self-attention. In this decoder, the query $\bq_t$ is processed using the key-value pairs $\bh_t$, with the reference point set to $\hat{\bp}^{{patch}}_t$. 
To limit the refinement to the local region, the offsets are constrained by the $S$ (stride) and mapped to the range $[-S, S]$ using a tanh activation.

In addition, we predict the visibility $\hat{\bv}_t$ and uncertainty $\hat{\bu}_t$, using visibility head $\Phi_{\text{vis}}$. We first decode the region around the predicted location $\hat{\bp}_t$ (Eq.~\ref{eq:patch_refinement}) using a deformable decoder layer. Then, we predict visibility and uncertainty by applying a linear layer to the decoded queries.
At training time, we define a prediction to be uncertain if the prediction error exceeds a threshold~($\delta_u = 8$ pixels) or if the point is occluded. 
During inference, we classify a point as visible if its probability exceeds a threshold $\delta_v$. Although we do not directly utilize uncertainty in our predictions during inference, we found predicting uncertainty to be beneficial for training.

\boldparagraph{Training} 
We train our model using the ground-truth trajectories $\bp_t \in \nR^{N \times 2}$ and visibility information $\bv_t \in \{0,1\}^{N}$. For patch classification, we apply cross-entropy loss based on the ground-truth class, patch $\bc^{patch}$.
For offset prediction $\hat{\bo}_t$, we minimize the $\ell_1$ distance between the predicted offset and the actual offset. We supervise the visibility $\hat{\bv}_t$ and uncertainty $\hat{\bu}_t$ using binary cross-entropy loss. Additionally, we supervise the uncertainties of the top-$k$ points, $\hat{\bu}^{top}_t$, at re-ranking. The total loss is a weighted combination of them:

\begin{ceqn}
\begin{equation}
    \begin{aligned}
        \cL = &~
        \lambda~ \underbrace{\left(\cL_\text{CE}\left(\bC_t,~\bc^{patch}\right) + 
                  \cL_\text{CE}\left(\bC^{dec}_t,~\bc^{patch}\right)\right)}_{\text{Patch Classification Loss}}  \cdot \bv_t\\
              &~ + \underbrace{\cL_{\ell_1}\left(\hat{\bo}_t,~\bo_t\right)}_{\text{ Offset Loss}} \cdot \bv_t 
              + \underbrace{\cL_\text{CE}(\hat{\bv}_t, \bv_t)}_{\text{Visibility Loss}} 
              + \underbrace{\cL_\text{CE}(\hat{\bu}_t, \bu_t)}_{\text{Uncertainty Loss}} 
              + \underbrace{\cL_\text{CE}(\hat{\bu}^{top}_t, \bu^{top}_t)}_{\text{Top-$k$ Uncertainty Loss}}
    \end{aligned}
\end{equation}
\end{ceqn}

\begin{figure}[t]
    \centering
    \includegraphics[width=0.8\linewidth]{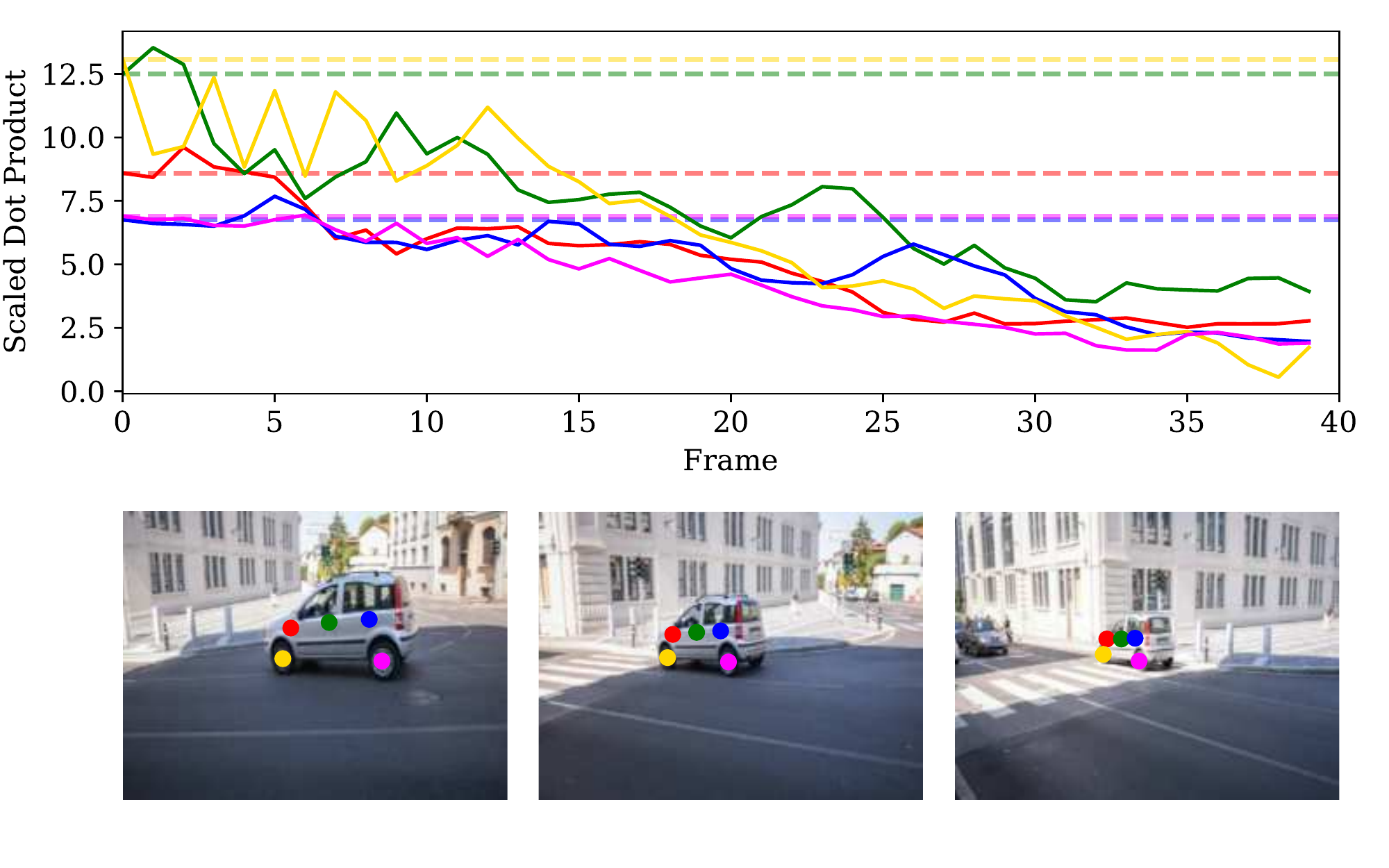}
        \caption{\textbf{Feature Drift.}
        For the tracks shown below (start, middle, and final frames), the plot above illustrates the decreasing similarity between the features of the initial query and its correspondences over time, with the initial similarity indicated by horizontal dashed lines.}
        \label{fig:track_on:drift}
\end{figure}

\noindent \textbf{Discussion}: 
Till this point, our model has exclusively considered relocating the queried points within the current frame. However, as the appearance of points consequently changes over time, the embedding similarity between the initial query point and future correspondences tends to decrease gradually (\figref{fig:track_on:drift}). This problem, known as feature drift, leads to inaccurate predictions, when solely relying on the feature similarity with the initial point.

\begin{figure}[t]
    \centering
    \includegraphics[width=\linewidth]{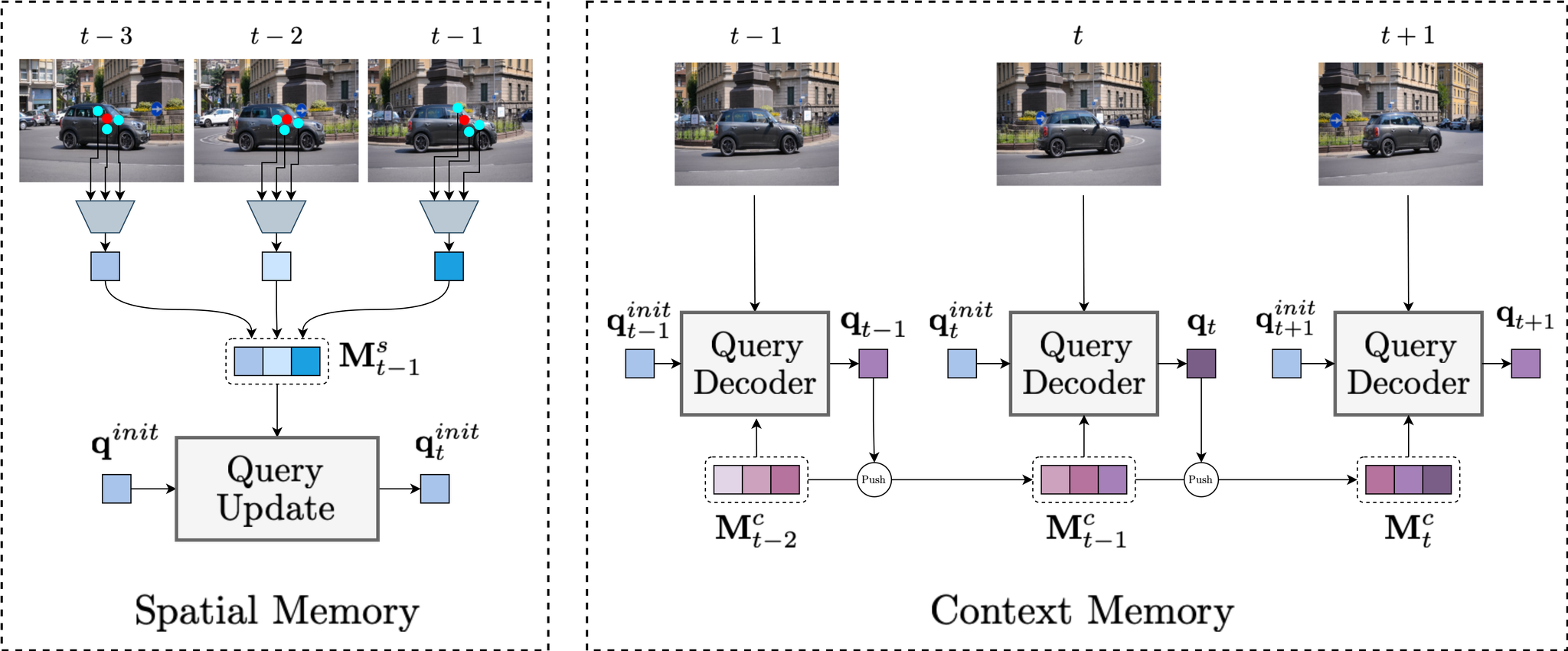}
        \caption{\textbf{Memory Modules.} Spatial memory $\bM_{t-1}^s$ (\textbf{left}) is used to update the initial query $\bq^{init}$ from the first frame to $\bq_t^{init}$ on the current frame. 
    The goal is to resolve feature drift by storing the content around the model’s predictions in previous frames. Context memory $\bM_{t-1}^c$ (\textbf{right}) is input to the query decoder which updates $\bq_t^{init}$ to $\bq_t$. It provides a broader view of the track’s history with appearance changes and occlusion status by storing the point’s embeddings from past frames.}
        \label{fig:track_on:memories}
\end{figure}

\subsection{Track-On with Memory}
\label{sec:track_on:full_model}

Here, we introduce two types of memories: \textbf{spatial memory} and \textbf{context memory}, as illustrated in~\figref{fig:track_on:memories}.
Spatial memory stores information around the predicted locations, allowing us to update the initial queries based on the latest predictions. Context memory preserves the track's history states by storing previously decoded queries, ensuring continuity over time and preventing inconsistencies. 
This design enables our model to effectively capture temporal progressions in long-term videos, 
while also adapting to changes in the target’s features to address feature drift. 

We store the past features for each of the $N$ queries independently, with up to $K$ embeddings per query in each memory module. Once fully filled, the earliest entry from the memory will be obsoleted as a new entry arrives, operating as a First-In First-Out~(FIFO) queue.

\subsubsection{Spatial Memory}
Here, we introduce the spatial memory module that stores fine-grained local information from previous frames, enabling continuous updates to the initial query points. This adaptation to appearance changes helps mitigate feature drift.

\boldparagraph{Memory Construction} 
We zero-initialize the memory, $\bM^s_0$, update its content with each frame. For the first frame, we make a prediction using initial query $\bq^{init}$ without memory. 

\boldparagraph{Memory Write ($\Phi_\text{\normalfont q-wr}$)} 
To update the memory with the new prediction, 
$\bM^s_{t-1} \rightarrow \bM^s_t$, 
we extract a feature vector around the predicted point $\hat{\bp}_t$ on the current feature map $\bff_t$, and add it to the memory:
\begin{ceqn}
\begin{equation}
\bM^s_{t} = [\bM^s_{t-1},~ \Phi_\text{\normalfont q-wr}\left([\bq^{init}, \bq_t],~ \bff_t,~ \hat{\bp}_t\right) ]
\end{equation}
\end{ceqn}
$\Phi_\text{q-wr}$ is a 3-layer deformable transformer decoder without self-attention, 
using the concatenated $\bq^{init}$ and $\bq_t$ as the query, attending a local neighborhood of predicted point for update. Utilizing deformable attention for the local summarization process helps prevent error propagation over time, as the query can flexibly select relevant features from any range.

\boldparagraph{Query Update ($\Phi_\text{\normalfont q-up}$)} 
In such scenario, before passing into the query decoder to estimate the correspondence, 
the initial query points first visit the spatial memory $\bM^s_{t-1}$ for an update:
\begin{ceqn}
\begin{equation}
\bq^{init}_t = \Phi_\text{q-up}\left(\bq^{init},~ \bM^s_{t-1} \right) = \bq^{init}  + \phi_\text{qqm}\left(\bq^{init},~ \phi_\text{mm}(\bM^s_{t-1} + \gamma^s)\right)
\end{equation}
\end{ceqn}
$\phi_\text{mm}$ is a transformer encoder layer that captures dependencies within the memory, and $\phi_\text{qqm}$ is a transformer decoder layer without initial self-attention, where $\bq^{init}$ attends to updated memory, followed by a linear layer, and $\gamma^s \in \nR^{K \times D}$ is learnable position embeddings. 
Instead of sequentially updating the query embeddings at each time step, 
\eg extracting $\bq_{t}^{init}$ using $\bq_{t-1}^{init}$, 
we update them with respect to the initial query $\bq^{{init}}$, 
conditioned on all previous predictions stored in the memory. 
This prevents error propagation by taking into account the entire history of predictions.

\subsubsection{Context Memory} 
In addition to spatial memory, we introduce a context memory that incorporates historical information of the queried points from a broader context, enabling the model to capture past occlusions and visual changes. Specifically, we store the decoded query features from previous time steps in context memory, $\bM^c_{t-1}$. We then integrate it by extending the query decoder with an additional transformer decoder layer without self-attention, where queries attend to memory with added learnable position embeddings~($\gamma^c \in \nR^{K \times D}$):
\begin{ceqn}
\begin{equation}
\bq_t =  \Phi_{\text{q-dec}} \left(\bq^{init}_t,~ \bff_{t},~ {\color{red} \bM^c_{t-1} +\gamma^c } \right)
\end{equation}
\end{ceqn}
Changes to the query decoder with memory are shown in red. For the writing operation, we add the most recent $\bq_t$ to $\bM^c_{t-1}$ and remove the oldest item, following the same procedure as in the spatial memory. Our experiments demonstrate that incorporating past content temporally with context memory enables more consistent tracking with additional benefits over spatial memory, especially in visibility prediction, since spatial memory focuses only on the regional content where the point is currently visible. 

\subsubsection{Inference-Time Memory Extension} 
Although the memory size $K$ is fixed at training time, 
the number of video frames at inference can be different from the training frame limit. 
To address this, we extend the memory size during inference by linearly interpolating the temporal positional embeddings, $\gamma^s$ and $\gamma^c$, to a larger size $K_i$. 
In particular, we train our model with memory size $K = 12$, and extend it to $K_i \in \{16, \dots, 96\}$ at inference time.

\section{Experiments}
\label{sec:track_on:exp}

\subsection{Experimental Setup}

\boldparagraph{Datasets}  
We evaluate our model on seven datasets with diverse characteristics. For both training and evaluation, we use TAP-Vid~\cite{Doersch2022NeurIPS}, consistent with prior work. Specifically, we train on TAP-Vid Kubric, a synthetic dataset containing 11K video sequences, each with 24 frames. For evaluation, we use three subsets from the TAP-Vid benchmark: \textbf{TAP-Vid DAVIS}, consisting of 30 real-world videos from the DAVIS dataset;  \textbf{TAP-Vid RGB-Stacking}, a synthetic dataset of 50 videos focused on robotic manipulation with mostly textureless objects; \textbf{TAP-Vid Kinetics}, which includes over 1,000 real-world videos from the Kinetics dataset.Beyond TAP-Vid, we evaluate generalization to the following datasets:  \textbf{RoboTAP}~\cite{Vecerik2023ICRA}, consisting of 265 real-world robotic sequences, each averaging over 250 frames;  \textbf{Dynamic Replica}~\cite{Karaev2023CVPR}, a benchmark for 3D reconstruction with 20 sequences of 300 frames;  \textbf{BADJA}~\cite{Biggs2019ACCV}, a dataset for animal joint tracking, containing 7 sequences with sparse trajectories; \textbf{Point Odyssey (PO)}~\cite{Zheng2023ICCV}, which contains 12 long-form videos of up to 4325 frames, designed to test extreme tracking persistence.

\boldparagraph{Evaluation Details}  
We follow the standard protocol of TAP-Vid benchmark by first downsampling the videos to $256 \times 256$. We evaluate models in the queried first protocol, which is the natural setting for causal tracking. In this mode, the first visible point in each trajectory serves as the query, and the goal is to track that point in subsequent frames.

\subsection{Results}

\subsubsection{TAP-Vid Benchmark}
As shown in \tabref{tab:track_on:sota_first}, we categorize models into online and offline settings. 
Offline models, with bidirectional information flow, use either a fixed-size window—where half of the window spans past frames and the other half future frames—or the entire video, granting access to any frame regardless of video length and providing a clear advantage. In contrast, online models process one frame at a time, enabling frame-by-frame inference. In the following discussion, we mainly focus on the setting using similar training set to ours, \ie the models without using real-world videos.

We evaluate tracking performance with the following metrics of TAP-Vid benchmark, as in Chapter~\ref{chapter:eval_fomo}: Occlusion Accuracy (OA), which measures the accuracy of visibility prediction; \deltaavg, the average proportion of visible points tracked within 1, 2, 4, 8, and 16 pixels; Average Jaccard (AJ), which jointly assesses visibility and localization precision. 

\begin{table}
    \centering
    \footnotesize
    \setlength{\tabcolsep}{3pt}
    \caption{\textbf{Quantitative Results on TAP-Vid Benchmark.} 
    This table shows results in comparison to the previous work on TAP-Vid under queried first setting, in terms of AJ, \deltaavg, and OA.
    The models are categorized into online and offline schemes, the former setting grants access to any frame regardless of video length, thus providing a clear advantage. While online models process one frame at a time, enable frame-by-frame inference. 
    For training datasets, Kub and Kub-L(ong), refer to the TAP-Vid Kubric dataset with 24-frame and 64-frame videos, respectively; and R indicates the inclusion of a large number of real-world videos, we highlight these models in \colorbox{lightgray}{gray}. MFT is a long-term optical flow method trained on a combination of Sintel~\cite{Butler2012ECCV}, FlyingThings~\cite{Mayer2016CVPR}, and Kubric datasets.}

    \begin{tabular}{l!{\vrule width -1pt}c!{\vrule width -1pt}c!{\vrule width -1pt}c!{\vrule width -1pt}c!{\vrule width -1pt}c!{\vrule width -1pt}c!{\vrule width -1pt}c!{\vrule width -1pt}c!{\vrule width -1pt}c!{\vrule width -1pt}c!{\vrule width -1pt}c}
        \toprule
        \multicolumn{1}{l}{\multirow{3}{*}{\textbf{Model}}} & \multicolumn{1}{c}{\multirow{3}{*}{Input}} & \multicolumn{1}{c}{\multirow{3}{*}{Train}} & \multicolumn{3}{c}{DAVIS} & \multicolumn{3}{c}{RGB-Stacking} & \multicolumn{3}{c}{Kinetics} \\
        \cmidrule(r){4-6} \cmidrule(r){7-9} \cmidrule(r){10-12}
         \multicolumn{3}{c}{} &  AJ \up &  \deltaavg \up & \multicolumn{1}{c}{OA \up} & AJ \up  & \deltaavg \up & \multicolumn{1}{c}{OA \up} & AJ \up  & \deltaavg \up & OA \up \\
         \midrule
         \textbf{Offline} & & & & & & & & & & & \\ 
         TAPIR & Video & Kub & 56.2 & 70.0 & 86.5 & 55.5 & 69.7 & 88.0 & 49.6 & 64.2 & 85.0 \\
         TAPTR & Window & Kub & 63.0 & 76.1 & \underline{91.1} & 60.8 & 76.2 & 87.0 & 49.0 & 64.4 & 85.2 \\
         TAPTRv2 & Window & Kub & 63.5 & 75.9 & \textbf{91.4} & 53.4 & 70.5 & 81.2 & 49.7 & 64.2 & 85.7 \\
         SpatialTracker & Window & Kub & 61.1 & 76.3 & 89.5 & 63.5 & 77.6 & 88.2 & 50.1 & 65.9 & 86.9 \\ 
         LocoTrack & Video & Kub & 62.9 & 75.3 & 87.2 & 69.7 & 83.2& 89.5 & 52.9 & \underline{66.8} & 85.3 \\ 
         CoTracker3 & Window & Kub-L & 64.5 & 76.7 & 89.7 & 71.1 & 81.9 & 90.3 & \textbf{54.1} & 66.6 & \underline{87.1} \\
         CoTracker3 & Video & Kub-L & 63.3 & 76.2 & 88.0 & \textbf{74.0} & \underline{84.9} & \underline{90.5} & 53.5 & 66.5 & 86.4 \\
         \rowcolor{lightgray}
         BootsTAPIR & Video & Kub + R & 61.4 & 73.6 & 88.7 & 70.8 & 83.0 & 89.9 & 54.6 & 68.4 & 86.5 \vspace{-1pt} \\
         \rowcolor{lightgray}
         CoTracker3 & Window & Kub-L + R & 63.8 & 76.3 & 90.2 & 71.7 & 83.6 & 91.1 & 55.8 & 68.5 & 88.3 \vspace{-1pt} \\
         \rowcolor{lightgray}
         CoTracker3 & Video & Kub-L + R & 64.4 & 76.9 & 91.2 & 74.3 & 85.2 & 92.4 & 54.7 & 67.8 & 87.4 \\
         \midrule
         \textbf{Online} &  & & & & & & & & \\  
         DynOMo & Frame & - & 45.8 & 63.1 & 81.1 & - & - & - & - & - & - \\
         MFT & Frame & SFK & 47.3 & 66.8 & 77.8 & - & - & - & 39.6 & 60.4 & 72.7 \\
         Online TAPIR & Frame & Kub & 56.7 & 70.2 & 85.7 & 67.7 & - & - & 51.5 & 64.4 & 85.2 \\
         Track-On (\textit{Ours}) & Frame & Kub & \textbf{65.0} & \textbf{78.0} & 90.8 & \underline{71.4} & \textbf{85.2} & \textbf{91.7} & \underline{53.9} & \textbf{67.3} & \textbf{87.8} \\
         
        \bottomrule
    \end{tabular}
    \label{tab:track_on:sota_first}
\end{table}

\boldparagraph{Comparison on DAVIS} Our model outperforms all existing online models across every evaluation metric, achieving an 8.3 AJ improvement over the closest competitor, Online TAPIR. Additionally, it surpasses all offline models in both AJ (65.0 \vs 64.5) and \deltaavg (78.0 \vs 76.7), outperforming even the concurrent CoTracker3, which was trained on longer videos (24 \vs 64 frames). Notably, our model also outperforms models fine-tuned on real-world videos by a significant margin. These results are particularly impressive because our model is an online approach, processing the video frame by frame, yet it exceeds the performance of offline models that process the entire video at once.

\boldparagraph{Comparison on RGB-Stacking} 
The dataset consists of long video sequences, with lengths of up to 250 frames, making it ideal for evaluating models' long-term processing capabilities. Our model surpasses Online TAPIR by 3.7 AJ and outperforms offline competitors, achieving improvements of 0.3 in \deltaavg and 1.2 in OA compared to CoTracker3, which utilizes video-level input. The results of offline models on this dataset highlight a significant limitation of the windowed inference approach, which struggles with long video sequences due to restricted temporal coverage. In contrast, models with full video input perform considerably better. By effectively extending the temporal span through our memory mechanisms, our model achieves comparable or superior performance on long videos using only frame-by-frame inputs, despite the inherent disadvantage of not having bidirectional connections across the entire video sequence.

\boldparagraph{Comparison on Kinetics} 
The dataset comprises a variety of long internet videos. 
Our model outperforms Online TAPIR across all metrics by a considerable margin, while also surpassing offline models in \deltaavg and OA. Specifically, it achieves a 0.5 improvement in \deltaavg over LocoTrack (with video inputs) and a 0.7 improvement in OA over CoTracker3 (with window inputs). 
Despite the significant difference in training data between CoTracker3 and our model, ours ranks second in AJ, with only a small gap of 0.2. Additionally, models fine-tuned on real-world data demonstrate superior performance, underscoring the potential benefits of training on large-scale real-world datasets, which seem particularly advantageous for datasets like Kinetics compared to others.

\begin{table}[h!]
    \centering
    \footnotesize
    \setlength{\tabcolsep}{3pt}
    \caption{\textbf{Quantitative Results on RoboTAP, Dynamic Replica, and BADJA} This table shows results in comparison to the previous work on RoboTAP, Dynamic Replica, and BADJA under queried first setting. The models are categorized into online and offline schemes, the former setting grants access to any frame regardless of video length, thus providing a clear advantage. While online models process one frame at a time, enable frame-by-frame inference. 
    For training datasets, Kub and Kub-L(ong), refer to the TAP-Vid Kubric dataset with 24-frame and 64-frame videos, respectively; and R indicates the inclusion of a large number of real-world videos, we highlight these models in \colorbox{lightgray}{gray}.} 
    \begin{tabular}{l!{\vrule width -1pt}c!{\vrule width -1pt}c!{\vrule width -1pt}c!{\vrule width -1pt}c!{\vrule width -1pt}c!{\vrule width -1pt}c!{\vrule width -1pt}c!{\vrule width -1pt}c!{\vrule width -1pt}c}
        \toprule
        \multicolumn{1}{l}{\multirow{3}{*}{\textbf{Model}}} & \multicolumn{1}{c}{\multirow{3}{*}{Input}} & \multicolumn{1}{c}{\multirow{3}{*}{Train}} & \multicolumn{3}{c}{RoboTAP} & \multicolumn{1}{c}{Dynamic Replica} & \multicolumn{2}{c}{BADJA} \\
        \cmidrule(r){4-6} \cmidrule(r){7-7} \cmidrule(r){8-10}
         \multicolumn{3}{c}{} &
         AJ \up &  \deltaavg \up & \multicolumn{1}{c}{OA \up} & \multicolumn{1}{c}{$\delta^{vis}$ \up} & $\delta^{seg}$ \up  & $\delta^{3px}$ \up &  \\
         \midrule
         \textbf{Offline} \\ 
         TAPIR & Video & Kub & 59.6 & 73.4 & 87.0 & 66.1 & 66.9 & 15.2 \\
         TAPTR & Window & Kub & 60.1 & 75.3 & 86.9 & 69.5 & 64.0 & 18.2 \\
         TAPTRv2 & Window & Kub & 60.9 & 74.6 & 87.7 & - & - & - \\
         SpatialTracker & Window & Kub & - & - & - & - & 69.2 & 17.1 \\
         LocoTrack & Video & Kub & 62.3 & 76.2 & 87.1 & 71.4 & - & - \\
         CoTracker3 & Window & Kub-L & 60.8 & 73.7 & 87.1 & 72.9 & - & - \\ 
         CoTracker3 & Video & Kub-L & 59.9 & 73.4 & 87.1 & 69.8 & - & - \\ 
         \rowcolor{lightgray}
         BootsTAPIR & Video & Kub + R & 64.9 & 80.1 & 86.3 & 69.0 & - & - \vspace{-1pt} \\
         \rowcolor{lightgray}
         CoTracker3 & Window & Kub-L + R & 66.4 & 78.8 & 90.8 & 73.3 & - & - \vspace{-1pt} \\
         \rowcolor{lightgray}
         CoTracker3 & Video & Kub-L + R & 64.7 & 78.0 & 89.4 & 72.2 & - & - \vspace{-1pt} \\
         
         \midrule
         \textbf{Online} \\  
         Online TAPIR & Frame & Kub & 59.1 & - & - & - & - & - \\
         Track-On (\textit{Ours}) & Frame & Kub & \textbf{63.5} & \textbf{76.4} & \textbf{89.4} & \textbf{73.6} & \textbf{71.9} & \textbf{20.2} \\
         
        \bottomrule
    \end{tabular}
    \label{tab:track_on:sota_first_second}
\end{table}

\subsubsection{RoboTAP} 
We evaluate our model on the RoboTAP dataset~\cite{Vecerik2023ICRA}, 
which consists of 265 real-world robotic sequences with an average length of over 250 frames, 
as shown in~\tabref{tab:track_on:sota_first_second}. We use the same metrics as the TAP-Vid benchmark: AJ, $\delta_\text{avg}$, and OA, with the memory size $K_i$ set to 48. Our model consistently surpasses existing online and offline models across all metrics. 
Specifically, in AJ and $\delta_\text{avg}$, our model outperforms the closest competitor, LocoTrack (which processes the entire video), by 1.2 and 0.2 points, respectively. Additionally, it exceeds the nearest competitor (TAPTRv2) in OA by 1.7 points. This demonstrates that our causal memory modules, which enable online tracking, are capable of effectively capturing the dynamics of long video sequences despite lacking bidirectional information flow across all frames. It is worth noting that this dataset, which features textureless objects, presents a significant challenge. 
Fine-tuning on real-world videos provides substantial improvements, as learning to track points on textureless objects is particularly difficult, as highlighted by models tuned on real-world datasets.

\subsubsection{Dynamic Replica} We compare to previous work on the Dynamic Replica dataset~\cite{Karaev2023CVPR}, a benchmark designed for 3D reconstruction with 20 sequences, each consisting of 300 frames, as shown in~\tabref{tab:track_on:sota_first_second}. Following prior work~\cite{Karaev2024ECCV}, we evaluate models using $\delta^{\text{vis}}$, consistent with the TAP-Vid benchmark. Unlike previous work, we do not report $\delta^{\text{occ}}$, as our model is not supervised for occluded points. The memory size is set to $K_i = 48$. Despite being an online model, our model outperforms offline competitors, including those trained on longer sequences (CoTracker3, 73.6 \vs 72.9) and versions fine-tuned on real-world videos (73.6 \vs 73.3). This highlights the robustness of our model, particularly in handling longer video sequences effectively.

\subsubsection{BADJA} 
We compare to previous work on the BADJA challenge~\cite{Biggs2019ACCV}, a dataset for animal joint tracking comprising 7 sequences, as shown in \tabref{tab:track_on:sota_first_second}. Two metrics are used for evaluation: $\delta^{\text{seg}}$, which measures the proportion of points within a threshold relative to the segmentation mask size (specifically, points within $0.2\sqrt{A}$, where $A$ is the area of the mask); and $\delta^{\text{3px}}$, the ratio of points tracked within a 3-pixel range. Given the dataset’s low FPS nature, we kept the memory size at the original value of 12. Our model achieves state-of-the-art results by a significant margin, with a 2.7-point improvement in $\delta^{\text{seg}}$ over SpatialTracker and a 2.0-point improvement in $\delta^{\text{3px}}$ over TAPTR. These results highlight the flexibility of our inference-time memory extension, enabling the model to adapt effectively to data with varying characteristics.

\begin{table}[t]
    \centering
    \small
    \caption{\textbf{Quantitative Results on PointOdyssey.} This table shows results in comparison to the previous work on PointOdyssey under queried first setting.} 
    \begin{tabular}{lcc cccc}
        \toprule
        \multicolumn{1}{l}{\multirow{3}{*}{\textbf{Model}}} & \multicolumn{1}{c}{\multirow{3}{*}{Input}} & \multicolumn{1}{c}{\multirow{3}{*}{Train}} & \multicolumn{4}{c}{PointOdyssey}  \\
        \cmidrule(r){4-7} 
         \multicolumn{3}{c}{} &
         $\delta^{vis}_{avg}$ \up &  $\delta^{all}_{avg}$ \up & MTE $\downarrow$ & Survival \up  \\
         \midrule
         TAP-Net & Frame & Kub & - & 23.8 & 92.0 & 17.0 \\
         TAP-Net & Frame & PO & - & 28.4 & 63.5 & 18.3 \\
         PIPs & Window & Kub & - & 16.5 & 147.5 & 32.9 \\
         PIPs & Window & PO & - & 27.3 & 64.0 & 42.3 \\
         PIPs++ & Window & PO & 32.4 & 29.0 & - & 47.0 \\
         CoTracker & Window & PO & \underline{32.7} & \underline{30.2} & - & \textbf{55.2} \\
         Track-On (\textit{Ours}) & Frame & Kub & \textbf{38.1} & \textbf{34.2} & \textbf{28.8} & \underline{49.5} \\
        \bottomrule
    \end{tabular}
    \label{tab:track_on:po}
\end{table}

\subsubsection{PointOdyssey} We evaluated our model, trained on TAP-Vid Kubric, on the PointOdyssey (PO)~\cite{Zheng2023ICCV} dataset, which consists of 12 long videos with thousands of frames (up to 4325). The results are presented in ~\tabref{tab:track_on:po}. We adopted four evaluation metrics proposed in Point Odyssey: $\delta^{vis}_{avg}$, which measures the $\delta_{avg}$ metric from the TAP-Vid benchmark for visible points; $\delta^{all}_{avg}$, which calculates $\delta_{avg}$ for all points, including both visible and occluded ones; MTE (Median Trajectory Error), computed for all points; and Survival Rate, defined as the average number of frames until tracking failure (set to 50 pixels). The memory size $K_i$ was set to 96. From the results, we observe that PIPs trained on Kubric achieves a $\delta^{vis}_{avg}$ of 16.5, while the same model trained on PO with a larger window size achieves 27.3 ($\sim$ 65\% improvement). Notably, CoTracker does not report the performance of its model trained on Kubric but instead reports results for a model trained with sequences of length 56 on PO. These findings highlight the importance of training on PO to achieve higher performance across models. Our model, trained on Kubric, outperforms CoTracker and PIPs++ trained on PO in both $\delta^{vis}_{avg}$ and $\delta^{all}_{avg}$. Interestingly, while training on PO is critical for other models to achieve strong performance, our model demonstrates robustness by surpassing them even when trained on a different data distribution. Moreover, despite not being explicitly supervised for occluded points, our model still achieves superior $\delta^{all}_{avg}$. In terms of the Survival Rate, our model falls behind CoTracker trained on PO, despite its superior $\delta$ metrics. This further emphasizes the importance of training on PO to excel in this specific metric.

\subsection{Ablation Study}

\begin{table}[b!]
    \centering
    \caption{\textbf{Model Components.} Removing individual components of our model without (inference-time memory extension)—namely, the re-ranking module ($\Phi_\text{rank}$), offset head ($\Phi_\text{off}$), and visibility head ($\Phi_\text{vis}$) one at a time. All metrics are higher-is-better.}
    \begin{tabular}{l | cc | cc c}
         \toprule
         \textbf{Model} & $\delta^{1px}$ & $\delta^{16px}$ & AJ & \deltaavg & OA \\
         \midrule
         Full Model (without IME) & 45.5 & 95.9 & 64.9 & 77.7 & 90.6 \\
         - No re-ranking ($\Phi_\text{re-rank}$) & 43.8 & 95.5 & 62.8 & 76.3 & 89.7\\
         - No offset head ($\Phi_\text{off}$) & 27.6 & 96.1 & 60.1 & 73.0 & 90.5 \\
         - No visibility head ($\Phi_\text{vis}$) & 45.4 & 96.1 & 64.0 & 77.4 & 90.6  \\
       \bottomrule
    \end{tabular}
    \label{tab:track_on:component_ablation}
\end{table}

\boldparagraph{Components} We conducted an experiment to examine the impact of each proposed component in the correspondence estimation section (\secref{sec:track_on:vanilla_model}),
we remove them one at a time while keeping other modules unchanged. 
First, we removed the re-ranking module $\Phi_\text{rank}$. 
Second, we removed the offset head $\Phi_\text{off}$, eliminating the calculation of additional offsets. Instead, we used the coarse prediction, \ie the selected patch center, as the final prediction. Lastly, we replaced the additional deformable attention layer in the visibility head $\Phi_\text{vis}$ with a 2-layer MLP. Note that, we do not apply inference-time memory extension to models in this comparison.

From the results in~\tabref{tab:track_on:component_ablation}, we can make the following observations: 
(i) The re-ranking module improves all metrics, notably increasing AJ by 2.1, as it introduces specialized queries for identifying correspondences. Errors larger than 16 pixels are also more frequent without it, showing its role in reducing large errors. 
(ii) The offset head is crucial for fine-grained predictions. While $\delta^{16px}$ values remain similar without the offset head, lower error thresholds (\ie less than 1 pixel) show a significant difference (45.5 \vs 27.6), highlighting the importance of predicted offsets for fine-grained localization. (iii) Replacing the deformable attention layer in $\Phi_\text{vis}$ with an MLP does not affect OA but reduces AJ. The deformable head ensures more consistent visibility predictions by conditioning them on accurate point predictions, leading to higher AJ. Despite this, OA remains robust even when an MLP is used for visibility prediction.

\begin{table}[b]
    \centering
    \caption{\textbf{Offset Head.} The effect of removing the offset head ($\Phi_\text{off}$) on models with varying strides. All metrics are higher-is-better.}
    \begin{tabular}{c c | ccc | ccc}
        \toprule
         $\Phi_\text{off}$ & Stride & $\delta^{2px}$ & $\delta^{4px}$ & $\delta^{8px}$ & AJ & \deltaavg & OA   \\
         \midrule
         \xmark & \multirow{2}{*}{8} & 37.4 & 79.0 & 91.1 & 51.3 & 62.9 & 91.0 \\
         \cmark &  & 66.1 & 84.0 & 91.7 & 62.5 & 75.8 & 90.6 \\
         \midrule
         \xmark & \multirow{2}{*}{4} & 64.3 & 84.4 & 92.4 & 60.1 & 73.0 & 90.5 \\
         \cmark & & 69.3 & 85.5 & 92.5 & 64.9 & 77.7 & 90.6 \\
       \bottomrule
    \end{tabular}
    \label{tab:track_on:offset_ablation}
\end{table}

\boldparagraph{Offset Head and Stride} The offset head is essential for refining patch classification outputs, enabling more precise localization. Specifically, the offset head allows for precision beyond the patch size $S$ (stride). In~\tabref{tab:track_on:offset_ablation}, we examine the impact of removing the offset head ($\Phi_\text{off}$) for two stride values, $S = 4$ and $S = 8$, without utilizing inference-time memory extension. For both values, the addition of the offset head significantly enhances AJ and \deltaavg by refining predictions within the local region. With stride 4, the offset head notably improves $\delta^{2px}$, while for stride 8, it improves both $\delta^{2px}$ and $\delta^{4px}$. This demonstrates that while patch classification offers coarse localization, the offset head provides further refinement, achieving pixel-level precision. 

Larger stride values risk losing important details necessary for accurate tracking. For instance, increasing the stride from 4 to 8 results in AJ drops of 12\% for TAPIR and 16\% for CoTracker, as reported in their ablation studies. However, our coarse-to-fine approach mitigates the negative effects of stride 8, leading to only a minimal decline of 4\%, highlighting the robustness of our model to larger stride values.

Note that the model with $S=8$ and no offset head (first row) has a higher occlusion accuracy (OA). A possible reason is the imbalance in the loss, where the visibility loss has a relatively higher impact compared to the model with an additional offset loss (second row), leading to improved occlusion accuracy.

\begin{table}[b!]
    \centering
    \caption{\textbf{Memory Components.} The effect of spatial memory ($\bM^s$), context memory ($\bM^c$), and inference-time memory extension (IME). All metrics are higher-is-better.}
    \begin{tabular}{c | c c c | ccc}
        \toprule
        \textbf{Model} & $\bM^s$ &  $\bM^c$ & IME  & AJ & \deltaavg  & OA \\
        \midrule
         A & \xmark & \xmark & \xmark & 52.0 & 67.6 & 78.1 \\
         B & \cmark & \xmark & \xmark & 63.5 & 77.0 & 89.0 \\
         C & \xmark & \cmark & \xmark & 64.3 & 77.8 & 90.3 \\
         D & \cmark & \cmark & \xmark & 64.9 & 77.7 & 90.6 \\
         E & \cmark & \cmark & \cmark & 65.0 & 78.0 & 90.8 \\
       \bottomrule
    \end{tabular}
    \label{tab:track_on:memory_ablation}
\end{table}

\boldparagraph{Memory Modules} 
To demonstrate the effectiveness of our proposed memory modules, 
we conduct an ablation study, as shown in~\tabref{tab:track_on:memory_ablation}. 
We start by evaluating the model without memory (Model-A), 
which corresponds to the vanilla model described in \secref{sec:track_on:vanilla_model}. 
As expected, due to the model's lack of temporal processing, Model-A performs poorly, particularly in OA. Introducing temporal information through either spatial memory (Model-B) or context memory (Model-C) leads to significant performance improvements.
Model-C, in particular, achieves higher OA by providing a more comprehensive view of the track's history, including occlusions. Combining both memory types (Model-D) further boosts performance, highlighting the complementary strengths of the two memory modules. Lastly, incorporating the memory extension at inference time yields slight improvements in all metrics, leading to an overall enhancement in performance.

\begin{figure}[t]
    \centering
    \includegraphics[width=0.7\linewidth]{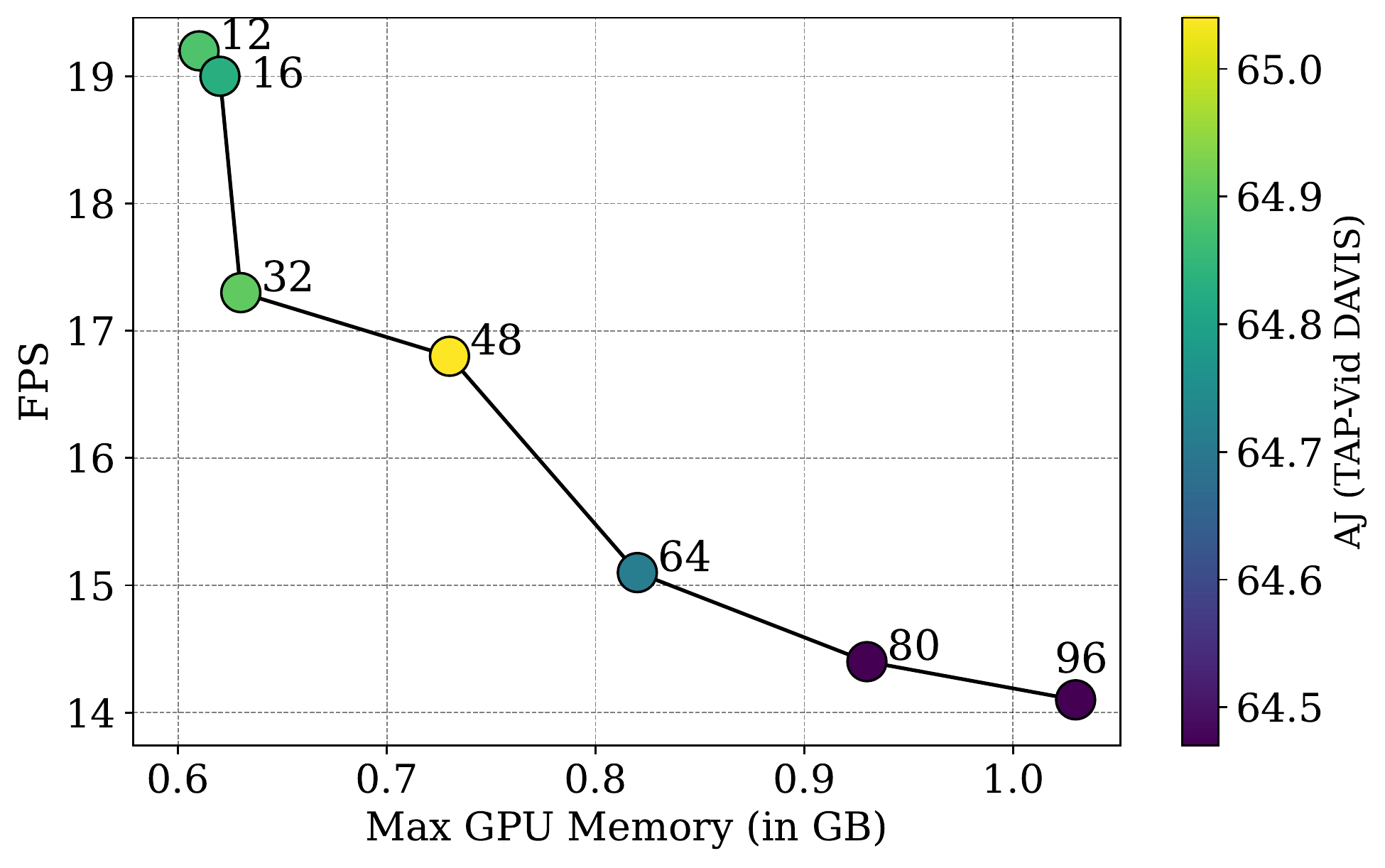}
    \caption{\textbf{Efficiency.} Inference speed (frames per second, FPS) vs. maximum GPU memory usage (in GB) where color represents the performance in AJ for different memory sizes (indicated near the nodes), while tracking approximately 400 points on the DAVIS dataset.}
    \label{fig:track_on:fps_vs_mem}
\end{figure}

\boldparagraph{Efficiency}
We plot the inference speed (frames per second, FPS), maximum GPU memory usage during video processing, and AJ performance on the TAP-Vid DAVIS dataset as a function of memory size $K_i$ (indicated near the plot nodes) in \figref{fig:track_on:fps_vs_mem}. The results are based on tracking approximately 400 points on a single NVIDIA A100 GPU. Unlike offline methods, our approach does not utilize temporal parallelization in the visual encoder, processing frames sequentially in an online setting. As the memory size $K$ increases, the model’s inference speed decreases due to the higher computational cost of temporal attention in memory operations, correspondingly increasing GPU memory usage. For instance, the FPS decreases from 19.2 with $K = 12$ to 16.8 with $K = 48$, and further down to 14.1 with $K= 96$.

Additionally, our model demonstrates high memory efficiency, with GPU memory usage ranging from 0.61 GB ($K = 12$) to a maximum of 1.03 GB ($K = 96$). At the default memory size of $K = 48$, where our model performs best on this dataset, it achieves 16.8 FPS with a maximum GPU memory usage of 0.73 GB. This highlights the efficiency of our frame-by-frame tracking approach, making it well-suited for consumer GPUs and real-time applications. Moreover, we observe that performance improves as the memory size increases up to $K = 48$, but declines beyond this point. This suggests that excessively large memory sizes can hurt performance by storing unnecessary information.

\begin{figure}[t]
    \centering
    \includegraphics[width=0.65\linewidth]{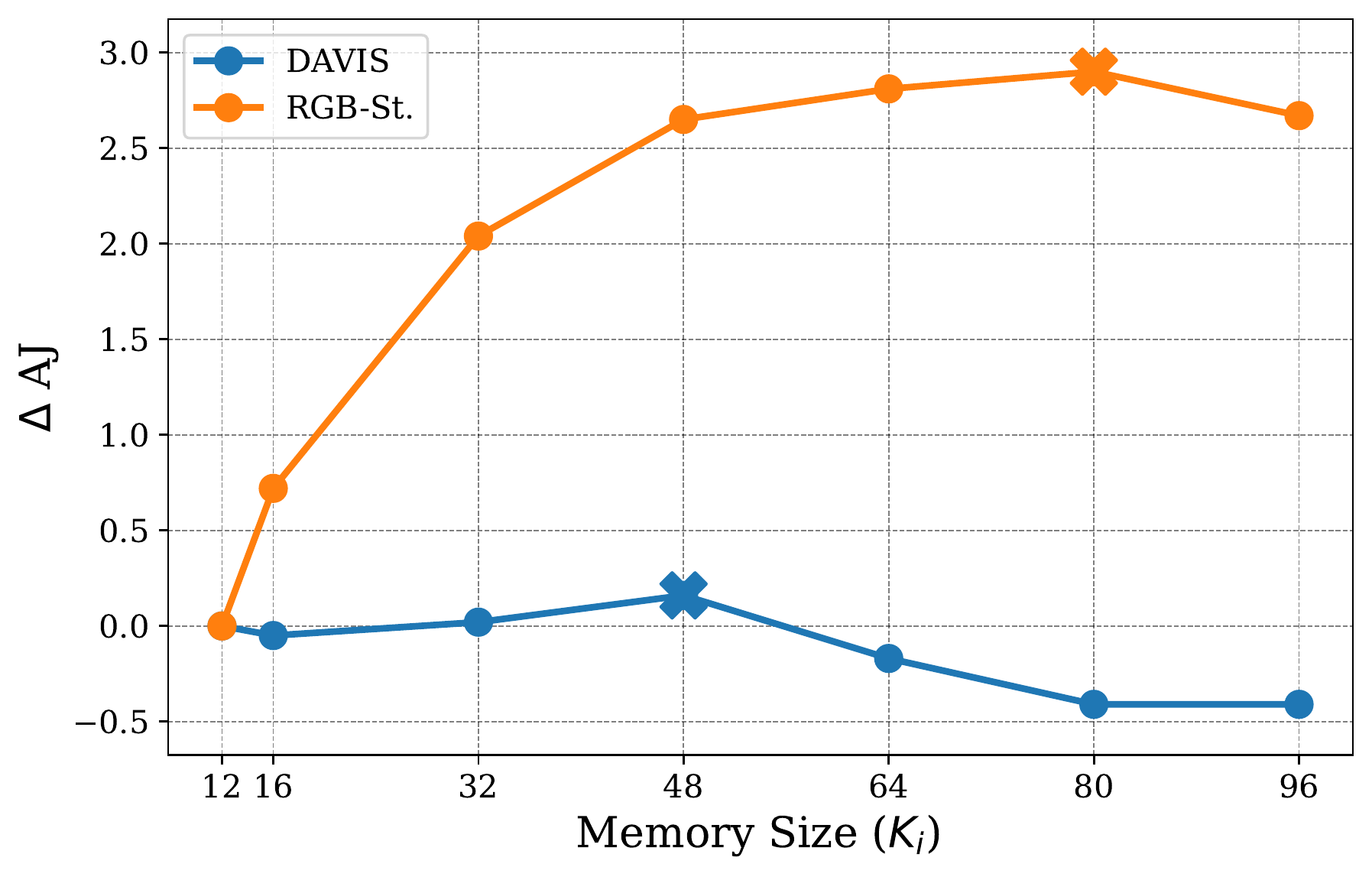}
    \caption{\textbf{Memory Size.} The effect of varying extended memory sizes during inference, on TAP-Vid DAVIS and TAP-Vid RGB-Stacking.}
    \label{fig:track_on:mem_vs_perf}
\end{figure}

\boldparagraph{Memory Size} We experimented with varying memory sizes, trained with $K=12$, and extended them to different values $K = \{16, 32, 48, 64, 80, 96\}$ during inference on TAP-Vid DAVIS and TAP-Vid RGB-Stacking, as shown in \figref{fig:track_on:mem_vs_perf}. The plot shows the change in AJ compared to the default training memory size of 12 after extension. Memory sizes reported in \tabref{tab:track_on:sota_first} are marked with crosses. For DAVIS ({\color{RoyalBlue} blue}), performance slightly increases up to a memory size of 48 ($64.88 \text{ AJ} \rightarrow 65.01 \text{ AJ}$) but declines beyond that, indicating that excessive memory can negatively impact the model. In contrast, for RGB-Stacking ({\color{orange} orange}), memory size plays a more critical role due to the disparity in video frame counts between training (24 frames) and inference (250 frames), as well as the high FPS nature of the dataset. Performance consistently improves up to $K = 80$, yielding a 2.9 AJ increase. These results highlight that, although the model is trained with a fixed and relatively small memory size, extending memory during inference is possible to adapt the varying characteristics of different datasets.

\boldparagraph{Spatial memory} To evaluate the effect of spatial memory in the presence of feature drift and inference-time memory extension, we conduct an experiment across different datasets using a model trained without spatial memory (Model-C in~\tabref{tab:track_on:memory_ablation}), as shown in~\tabref{tab:track_on:m_s_comp}. The results indicate that spatial memory consistently improves AJ across four datasets: DAVIS, RGB-Stacking, Kinetics, and RoboTAP. The impact is particularly notable for RGB-Stacking (+1.2 AJ) and RoboTAP (+1.4 AJ), where objects are less descriptive and often textureless, as both datasets originate from robotics scenarios. This suggests that spatial memory, which retains information around the local region of previous predictions, helps mitigate drift and enhances generalization across different scene characteristics.

\begin{table}[t]
    \centering
    \small
    \caption{\textbf{Spatial Memory.} Comparison of the model’s performance with and without spatial memory ($\bM^s$), evaluated using the AJ metric across different datasets, with inference-time memory extension (IME) applied.}
    \begin{tabular}{l | c c c c }
        \toprule
        \textbf{Model} & DAVIS & RGB-Stacking & Kinetics & RoboTAP \\
        \midrule
        Full Model & 65.0 & 71.4 & 53.9 & 63.5 \\
        - Without Spatial Memory ($\bM^s$) & 64.6 & 70.2 & 53.3 & 62.1 \\
        \bottomrule
        \end{tabular}
    \label{tab:track_on:m_s_comp}
\end{table}

Additionally, to directly assess the impact of spatial memory~(\secref{sec:track_on:full_model}) in mitigating feature drift, we conducted an analysis comparing the tracking performance of the initial feature sampled from the query frame, $\bq^{init}$, with the query feature updated using spatial memory at frame $t$, denoted as $\bq^{init}_t$. For this evaluation, we introduced the new metric of similarity ratio score ($s_{sr}$), which measures how well the updated query features align with the feature at the target point compared to the initial query.

Ideally, $\bq^{init}_t$ should provide a better starting point for detecting correspondences compared to $\bq^{init}$, particularly when the object’s appearance changes significantly. To assess whether $\bq^{init}_t$ is more similar to the feature at the ground-truth correspondence location than $\bq^{init}$, we calculate the ratio of their similarity to ground-truth, as a way of quantifying the increase in the similarity after the update:
\begin{ceqn}
\begin{equation}
    s_{sr} (t) = \dfrac{\bq^{init}_t \cdot \text{sample}(\bff_t, ~\bp_t)}{\bq^{init} \cdot \text{sample}(\bff_t, ~\bp_t)}
\end{equation}
\end{ceqn}

\begin{figure}[!b]
    \centering
    \begin{minipage}{.5\linewidth}
        \centering
        \includegraphics[width=\linewidth]{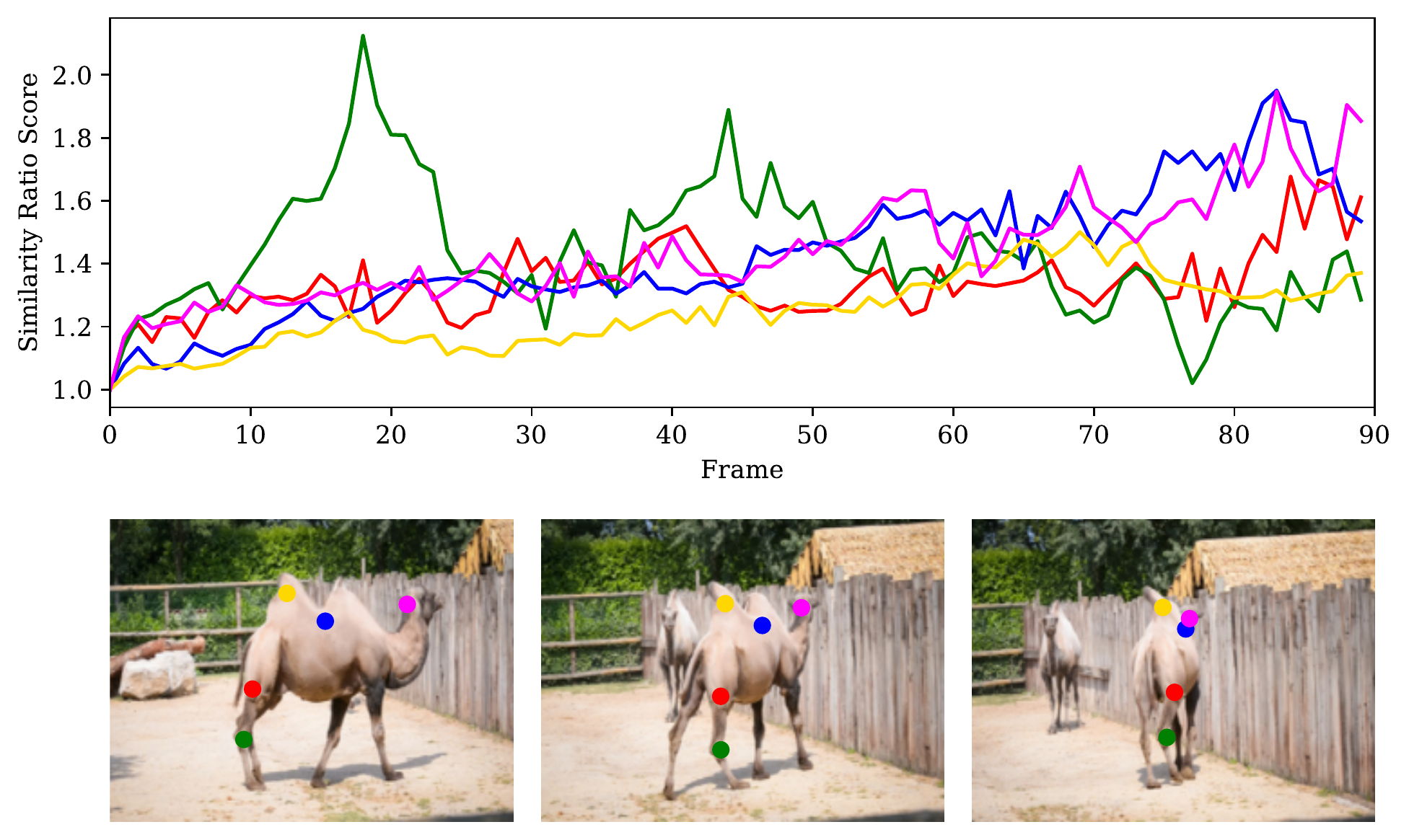}
        \label{fig:sub1}
    \end{minipage}%
    \begin{minipage}{.5\linewidth}
        \centering
        \includegraphics[width=\linewidth]{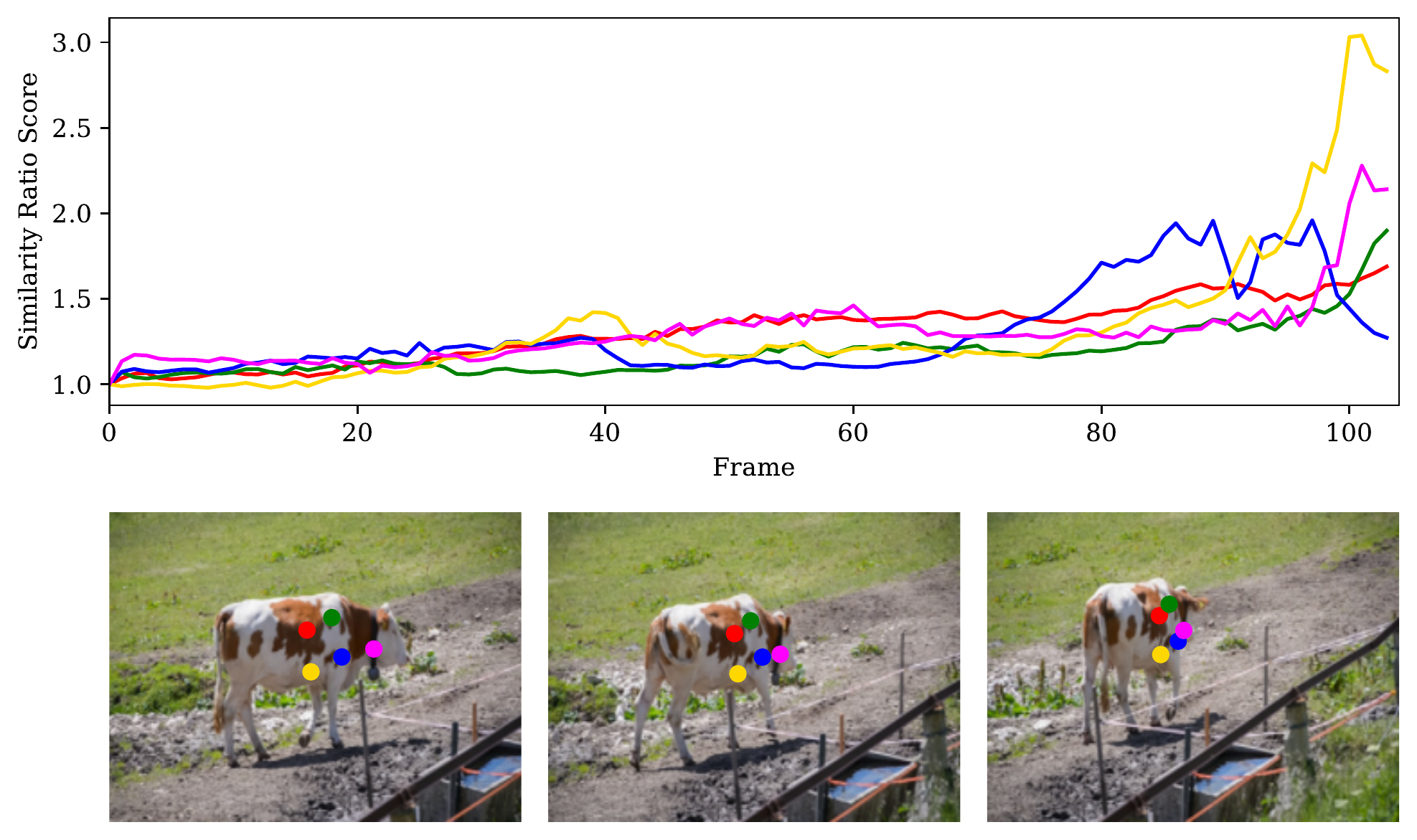}
        \label{fig:sub2}
    \end{minipage}
    \caption{\textbf{Similarity Ratio Score.} The similarity ratio score $s_{sr} > 1$ over frames for different tracks, demonstrates increased similarity with ground-truth location on the target frame when utilizing spatial memory.}
    \label{fig:track_on:sim_ratio}
\end{figure}

Here, $\bp_t$ represents the location of the ground-truth correspondence point, and $\bff_t$ is the feature map of the target frame. On the DAVIS dataset, we calculated $s_{sr}$ for visible points, achieving a score of 1.24, indicating that spatial memory introduces a 24\% increase in similarity compared to the initial feature. In~\figref{fig:track_on:sim_ratio}, we visualize the similarity scores for different tracks over time for two videos from the DAVIS dataset. The plot highlights that the similarity increases more significantly toward the end of the video, where appearance changes are more severe. Moreover, the score is consistently greater than 1, showing that $\bq^{init}_t$ always provides better initialization than $\bq^{init}$ in these videos.

\chapter{Conclusion}
\label{chapter:conc}
\section{Discussion}

In this thesis, we investigated the task of long-term point tracking with a focus on online settings, where the model must process video frames sequentially and cannot rely on future information. We approached this challenge from two complementary perspectives: evaluating the geometric capabilities of visual foundation models for point tracking, and designing a purpose-built transformer architecture that enables causal, efficient, and scalable tracking over long video sequences.

In Chapter~\ref{chapter:eval_fomo}, we analyzed whether FoMos trained on large-scale data can support point tracking through simple geometric similarity. Our findings revealed that certain FoMos exhibit strong geometric awareness, especially Stable Diffusion in zero-shot settings, and DINOv2 under lightweight adaptation. In particular, we showed that DINOv2 can match and even surpass the performance of fully supervised models, despite using fewer parameters and significantly less task-specific supervision. This demonstrates that these models encode meaningful correspondence priors and can serve as effective backbones for temporal reasoning tasks.

However, our evaluation also highlighted a key gap: while FoMos can localize points accurately across individual frames, they lack the temporal modeling capabilities required for consistent long-term tracking, especially under occlusion and appearance changes. This motivated the second part of the thesis, presented in Chapter~\ref{chapter:track_on}, where we introduced \textbf{Track-On}, a transformer-based model explicitly designed for online point tracking.

Track-On treats each point as a query token in a transformer decoder and estimates correspondences via patch classification followed by local offset refinement. To handle temporal consistency without requiring full video access, the model maintains two complementary memory modules: a spatial memory that stores localized features around previous predictions to reduce feature drift, and a context memory that aggregates the trajectory’s history to improve robustness and visibility estimation. This design enables the model to operate in a fully causal setting, tracking points frame-by-frame, while still capturing long-term dependencies.

Through extensive experiments on seven datasets, we demonstrated that Track-On achieves state-of-the-art performance among online models. It also competes closely with, and in some cases surpasses, offline models that have access to full videos or sliding windows of frames. We also show that Track-On is highly efficient, requiring minimal GPU memory and enabling fast inference, making it suitable for deployment in real-time applications.

Our ablation studies further validate the effectiveness of each component in our architecture. The patch classification and re-ranking modules improve robustness in coarse matching, while the offset head significantly enhances fine-grained localization. The memory modules play a crucial role in maintaining temporal coherence, particularly in long sequences. We also demonstrate that inference-time memory extension allows the model to scale beyond the fixed training window, adapting to varying sequence lengths without retraining.

In summary, this thesis presents two complementary contributions toward the goal of long-term online point tracking. First, we show that foundation models offer strong geometric cues that can be exploited with minimal supervision. Second, we propose a transformer-based architecture that translates these insights into an efficient and accurate online tracking model. Together, these components provide a practical and scalable solution to the problem of causal motion understanding in videos.

\section{Limitations and Future Work} 
A key limitation of this work, shared by most point tracking approaches, is the domain gap between training and test data. Current models are primarily trained on synthetic datasets such as TAP-Vid Kubric~\cite{Doersch2022NeurIPS}, where dense ground-truth correspondences can be generated automatically. In contrast, real-world datasets remain scarce and limited in diversity, as collecting ground-truth 2D trajectories across long videos is extremely labor-intensive and difficult to scale. As a result, available real-world datasets, such as TAP-Vid DAVIS or RoboTAP, are typically small in size and only used for benchmarking, not for training.

To address this, recent efforts like BootsTAP~\cite{Doersch2024ARXIV} and CoTracker3~\cite{Karaev2024ARXIV} explore pseudo-labeling strategies, bootstrapping real-world videos using predictions from existing models. While these methods offer improvements, they inherit the limitations of the teacher models. For instance, CoTracker3 is trained on pseudo-labels that are direct outputs of TAPIR and CoTracker, sharing the biases and failure modes of those models.

Emerging large-scale datasets such as Stereo4D~\cite{Jin2025CVPR} and DynPose-100K~\cite{Rockwell2025CVPR} open new possibilities for bridging this gap. These datasets use a combination of 3D scene reconstruction tools, like structure-from-motion, stereo depth estimation, and feature matching, to annotate massive collections of real-world internet videos with accurate camera poses and dynamic 3D point trajectories. For example, DynPose-100K achieves robust camera pose estimation for 100K dynamic videos using a fusion of advanced masking, tracking, and global bundle adjustment.

One promising direction is to leverage these reconstructions as a basis for extracting high-quality 2D point tracks from real-world 4D scenes. By projecting the stable 3D trajectories back to 2D views across time, we can construct a dataset of real-world point correspondences that covers a wide range of scene types, motions, and camera behaviors. Such a dataset would enable supervised training or self-training of point trackers with significantly improved realism and diversity, helping to close the domain gap between synthetic and real-world data. If adopted widely, this approach could benefit the broader point tracking community by establishing a scalable pipeline for collecting reliable training annotations from casually captured internet video.

Another important research direction involves improving point tracker design by leveraging scene-level geometry estimation. Recent models such as DUST3R~\cite{Wang2024CVPR}, VGGT~\cite{Wang2025CVPR} and MonST3R~\cite{Zhang2025ICLR} predict dense 3D structure from images. Particularly, MonST3R generalizes to dynamic scenes; and produces per-frame dynamic point clouds and camera poses. This opens up a new tracking formulation: instead of relying solely on 2D appearance features, we can lift the query point into the 3D space, track its motion in the world coordinate system, and reproject it into future frames using the estimated camera motion. This geometric perspective has the potential to resolve ambiguities in textureless regions, improve robustness to occlusion, and ensure physical consistency in the predicted trajectories. Integrating such geometric priors with our transformer-based tracking architecture is an exciting next step toward accurate and causally consistent tracking in the wild.

\bibliographystyle{apalike}
\bibliography{bibliography_long, references}

\end{document}